\documentclass{article}

 \usepackage[preprint]{neurips_2026}

\usepackage[utf8]{inputenc} % allow utf-8 input
\usepackage[T1]{fontenc}    % use 8-bit T1 fonts
\usepackage{hyperref}       % hyperlinks
\usepackage{url}            % simple URL typesetting
\usepackage{booktabs}       % professional-quality tables
\usepackage{amsfonts}       % blackboard math symbols
\usepackage{nicefrac}       % compact symbols for 1/2, etc.
\usepackage{microtype}      % microtypography
\usepackage{xcolor}         % colors

\usepackage{subcaption}
\usepackage{amsmath,amsthm}
\usepackage[table]{xcolor}
\usepackage{pgf}
\usepackage{multirow}
\usepackage{algorithm}

\usepackage{tikz}
\usetikzlibrary{positioning, shapes.geometric, arrows.meta}

\tikzset{
  person/.style={draw, circle, inner sep=2pt, minimum size=16pt},
  queryperson/.style={draw, circle, fill=red!10, inner sep=2pt, minimum size=16pt},
  edgelabel/.style={font=\ttfamily\footnotesize, fill=white, inner sep=1pt},
  >={Stealth},
  every node/.style={font=\footnotesize}
}
\colorlet{EdgeRed}{red} % your “pink” highlight

\usepackage[most]{tcolorbox}

\usepackage{wrapfig,framed}

\usepackage{listings}

\newcommand{\heat}[1]{%
    \pgfmathsetmacro{\opacity}{#1*60}% Scale 0-1 to 0-100 for xcolor
    \edef\myeq{\noexpand\cellcolor{blue!\opacity}}% Create the color command
    \myeq #1% Render the cell color and the value
}

\newtheorem{example}{Example}

\title{Project Auto-World: Towards Automated Benchmarking of Neural Relational Reasoners}

\author{%
Anirban Das\textsuperscript{*,1}~~~
Joanne Boisson\textsuperscript{*,1}~~~
Irtaza Khalid\textsuperscript{*,1}~~~
Sumita Garai\textsuperscript{2}~~~
Steven Schockaert\textsuperscript{1}
\\
\textsuperscript{1}Cardiff University \quad
\textsuperscript{2}University of Pennsylvania \quad
\\
\texttt{\{dasa8,boissonjc,khalidmi,schockaerts1\}@cardiff.ac.uk} \\ \texttt{sumita.garai@pennmedicine.upenn.edu}
}

\begin{document}
\renewcommand{\thefootnote}{\fnsymbol{footnote}}
\footnotetext[1]{co-first authors \\ Code is available at ~\url{https://github.com/autoworldrules/auto-world-rules/} }
\renewcommand{\thefootnote}{\arabic{footnote}}
\maketitle

\begin{abstract}
Reasoning about relational structures remains a significant challenge for neural models, particularly when they must systematically apply learned knowledge to problem instances that are harder  than those seen in training. Progress is hampered by the difficulty of evaluating such generalization, since \emph{a priori}, it is rarely clear what makes an instance hard.  We study how this issue can be addressed by using large language models (LLMs) to automate benchmark generation, learning to produce increasingly challenging instances in an end-to-end manner. Concretely, given a world parametrized by  Datalog rules, and an Edge Transformer as the reasoning evaluator, we use LLM-driven evolutionary search (based on FunSearch) and autonomous agentic search to discover sampling functions that yield  hard problem instances. We also show that the Edge Transformer can be improved using this data such that it generalizes well to further data perturbations. 
Finally, we show that the same machinery can be applied to novel worlds proposed by LLMs, opening the door to autonomous research on neural relational reasoning.
\end{abstract}

\section{Introduction}
The problem of inferring relationships between entities in a knowledge graph (KG) has received widespread attention \citep{DBLP:conf/nips/BordesUGWY13,DBLP:conf/ijcai/MeilickeCRS19,zhu2021neural}.
Despite the vast literature on this topic, however, most approaches are not capable of \textit{systematic} reasoning \citep{Sinha2019CLUTRR,nora}, i.e.\ they are not capable of applying the learned regularities in a systematic way to solve problems that are harder than those on which they were trained. This presents a stark contrast with classical approaches, such as Inductive Logic Programming \citep{DBLP:journals/ml/CropperDEM22}, where learned rules can be chained to make arbitrarily complex predictions. This observation has sparked a line of work that is focused on systematic neural relational reasoning \citep{DBLP:conf/icml/Minervini0SGR20,edge-transformer,r5,cheng2023neural,khalid2025systematic}.

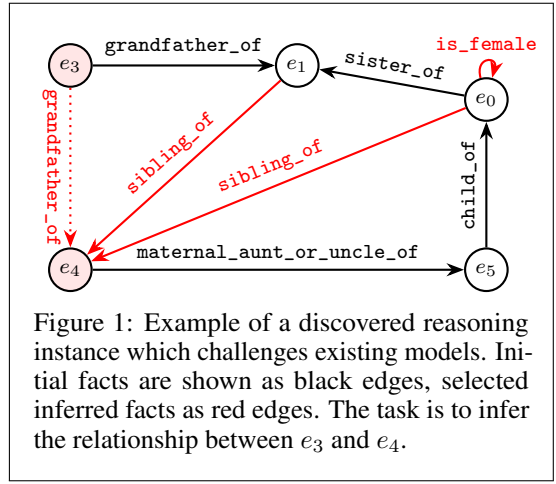
\begin{wrapfigure}{r}{205pt}
% \begin{subfigure}[b]{0.5\textwidth}
\begin{framed}
\begin{tikzpicture}[thick,->,x=1cm,y=0.9cm,baseline=(current bounding box.north)]
  % Nodes
  \node[queryperson] (e3) at (-2,3) {$e_3$};
  \node[person] (e1)  at (1,3) {$e_1$};
  \node[queryperson] (e4)   at (-2,0) {$e_4$};
  \node[person] (e5)  at (3.5,0)    {$e_5$};
  \node[person] (e0)  at (3.5,2.5) {$e_0$};

  % Edges
  \draw[EdgeRed] (e0) to[loop above, looseness=5]
        node[above]{\texttt{is\_female}} (e0);
  \draw[EdgeRed] (e1) -- (e4) node[midway,sloped, above]{\texttt{sibling\_of}};  
  \draw[EdgeRed] (e0) -- (e4) node[midway,sloped, above]{\texttt{sibling\_of}};  
  \draw[EdgeRed, dotted] (e3) -- (e4) node[midway,sloped, below]{\texttt{grandfather\_of}};  
  \draw (e5) -- (e0) node[midway,sloped, above]{\texttt{child\_of}};    
  \draw (e0) -- (e1) node[midway, sloped,above]{\texttt{sister\_of}};
  % \draw[densely dotted, bend left=30] (sam2) to
  %      node[pos=0.55, above, edgelabel]{\texttt{paternal\_grandma\_of}} (joe2);
  \draw (e3) -- (e1) node[midway, sloped, above]{\texttt{grandfather\_of}};
  \draw (e4) -- (e5) node[midway, sloped, above]{\texttt{maternal\_aunt\_or\_uncle\_of}};
\end{tikzpicture}
% \end{subfigure}
% \hfill
%     \begin{subfigure}[b]{0.5\textwidth}

% \caption{\label{fig1b}}
%     \end{subfigure}
\caption{Example of a discovered reasoning instance which challenges existing models. Initial facts are shown as black edges, selected inferred facts as red edges. The task is to infer the relationship between $e_3$ and $e_4$.
%We first need to derive that $e_0$ is female. This allows us to infer $\textit{mother\_of}(e_0,e_5)$ and then $\textit{sibling\_of}(e_0,e_4)$ and $\textit{sibling\_of}(e_1,e_4)$. Finally, we obtain $\textit{grandfather\_of}(e_3,e_4)$. 
\label{fig1a}}
\end{framed}
\end{wrapfigure}

To evaluate the systematic reasoning capabilities of these models, we need to create training/test splits in which the test examples are more difficult than the training instances, \emph{in some sense}. Crucially, however, this requires that we know what makes problem instances hard. Previous work has primarily considered the number of inference steps for this purpose \citep{Sinha2019CLUTRR}. 
However, the number of inference steps is not the only metric of difficulty that makes sense. For instance, \cite{khalid2025systematic} focused on problems where the answer can only be derived by combining multiple relational paths, finding that the number of such paths has a strong impact on problem difficulty. As another example, \cite{nora} found that existing models struggle when reasoning involves edges that are not on any path between the considered head and tail entities. 
Accordingly, they suggested using the number of such off-path edges as a difficulty metric.
This creates a conundrum: when we are evaluating reasoning models, we do not know \emph{a priori} what makes problem instances hard for these models. We can evaluate them w.r.t.\ known difficulty metrics, but these metrics may have blind spots, i.e.\ strong performance on the resulting training/test splits might mask their real limitations. Figure \ref{fig1a} illustrates this with an example that is hard for current models, despite not being identified as such by existing metrics (we will circle back to this example in the experiments section to discuss why we think it is difficult in a way that is not explained by known metrics).
 Furthermore, generating problem instances which are hard w.r.t.\ a given difficulty metric can be challenging in itself. The most common strategy is to create problem instances by randomly sampling KGs and then masking some edges, but the conditions that make reasoning hard are often unlikely to arise in these random graphs. While some authors have explored more structured sampling strategies for this reason \citep{Cohen2019GraphLog,khalid2025systematic}, such strategies cannot guarantee that problem instances have the ``right'' structure, given that we do not even know in advance what structures are needed to make reasoning challenging.

\textbf{Contributions.} 
%In this paper, 
Inspired by the successes of LLM-based approaches to scientific discovery \citep{DBLP:journals/corr/abs-2506-13131}, notably FunSearch \citep{funsearch}, we ask whether LLMs can automate the evaluation of relational reasoning models. 

We adapt FunSearch \citep{funsearch} into an evolutionary loop in which the LLM proposes Python \emph{priority functions} that drive the sampling of candidate problem instances, using the evaluator's failure rate as fitness. This 
%end-to-end evolutionary loop produces
leads to self-improving samplers that generate progressively harder instances. %at fixed graph size on a NoRA-style world where the resulting priority functions expose structural properties of hard instances that are not captured by existing difficulty metrics.
%such as inference depth, path multiplicity, or off-path entity count. 
Our analysis shows that the difficulty of the sampled problems cannot be (fully) explained by known difficulty metrics. We also show that surprisingly strong samplers can be obtained by prompting Claude Opus 4.6 in a dialogue setting.
Further, we find that augmenting the Edge Transformer's training data with instances generated by the learned samplers yields a stronger evaluator (\textsc{SuperET}) that largely resists further evolutionary perturbations within the FunSearch loop. However, replacing this loop with an end-to-end \texttt{autoresearch} style \citep{autoresearch} agent leads to samplers that can still challenge the \textsc{SuperET}. 
% recovers essentially the same priority functions; and we find that some of the discovered functions also transfer across reasoning worlds (NoRA $\rightarrow$ Iron Coast). 
% Finally, we show that the same machinery can be applied to novel worlds proposed by LMs (Iron Coast), opening the door to autonomous research on neural relational reasoning.

%******************************************************
\section{Background}
% Before discussing our LLM-based sampling strategies in Section \ref{secMethodology}, we recall some background on the problem of neural relational reasoning.

\paragraph{Knowledge representation}
A KG $\mathcal{G}$ is a set of triples $\mathcal{G}\subseteq \mathcal{E}\times\mathcal{R}\times\mathcal{E}$, where $\mathcal{E}$ is a set of entities and $\mathcal{R}$ is a set of relations. We can think of KGs as labelled graphs. For instance, if $\mathcal{G}$ contains a triple $(e,r,f)$, we say that there is an edge of type $r$ between $e$ and $f$. 
% Similarly, we say that there is a path of type $r_1;...;r_k$ between $e$ and $f$ if $\mathcal{G}$ contains triples of the form $(e,r_1,e_1),(e_1,r_2,e_2),...,(e_{k-1},r_k,f)$. 
At the same time, we can also think of KGs as sets of logical assertions. For example, the triple $(\textit{bob},\textit{father\_of},\textit{alice})$ can be viewed as the assertion that \textit{bob} is the father of \textit{alice}. 
% We use the notation $\textit{father\_of}(\textit{bob},\textit{alice})$ to denote such an assertion. 
Abusing notation, we will sometimes view KGs as sets of relational facts $r(e,f)$ rather than sets of triples $(e,r,f)$.
%
% \paragraph{Datalog rules} 
We rely on \textbf{Datalog rules} to encode the knowledge that allows us to reason about KGs.\footnote{Further details on Datalog can be found in Appendix \ref{appDetailsDatalog}} An example of a Datalog rule is as follows:
$$
\textit{father\_of}(X,Z) \leftarrow \textit{father\_of}(X,Y) \wedge \textit{brother\_of}(Y,Z)
$$
This rule encodes that whenever some entity $X$ is known to be the father of $Y$ and $Y$ is known to be the brother of $Z$, we can infer that $X$ is the father of $Z$. Given a KG $\mathcal{G}$ and a set of Datalog rules $\mathcal{P}$, we write $\mathcal{G}\cup\mathcal{P} \models r(e,f)$ to denote that $r(e,f)$ can be inferred from $\mathcal{G}$ using the rules in $\mathcal{P}$. For instance, in the example from Figure \ref{fig1a}, we can infer $\textit{grandfather\_of}(e_3,e_4)$ from the given KG and the rules from NoRA \citep{nora}. In this context, we often refer to KGs $\mathcal{G}$ as \textbf{worlds}. The rules in $\mathcal{P}$ capture the regularities that hold in these worlds; they are therefore sometimes referred to as \textbf{world rules}. In addition, we often also have access to a set of \textbf{constraints} $\mathcal{C}$. Each constraint corresponds to a conjunction of literals that can never be jointly satisfied (e.g.\ we cannot have $\textit{mother\_of}(X,Y)\wedge \textit{father\_of}(X,Y)$). 
% They are typically encoded as Datalog rules with an empty rule head:
% $$
% \leftarrow \textit{belongs\_to\_group}(Y, \textit{male}) \wedge \textit{belongs\_to\_group}(Y, \textit{female})
% $$
In the following, when generating worlds, we always guarantee that the constraints in $\mathcal{C}$ are satisfied, to prevent them from being non-nonsensical. %although the constraints themselves cannot be used for inferring knowledge. 

\paragraph{Difficulty metrics}
Current strategies for evaluating the systematic generalization abilities of neural reasoning models rely on metrics for assessing the difficulty of inferences of the form $\mathcal{G}\cup\mathcal{P}\models r(e,f)$. The most commonly used metric is the \textbf{inference depth (D)}, which is defined as the number of inference steps (i.e.\ rule applications) that are needed to infer $r(e,f)$. 
While standard Graph Neural Networks (GNNs) and Recurrent Neural Networks (RNNs) struggle to generalize to longer paths, experiments on benchmarks such as CLUTRR suggest that state-of-the-art models are rather robust in this respect \citep{DBLP:conf/iclr/ChengAS23,khalid2025systematic}.
\cite{nora} introduced two additional difficulty metrics, both of which capture the intuition that off-path reasoning is harder than path-based reasoning. The \textbf{Off-Path Edge Count (OPEC)} is defined as the number of triples that are used to infer $r(e,f)$ which do not correspond to an edge of a path from $e$ to $f$. 
The \textbf{Backtrack Load (BL)} instead measures the amount of off-path reasoning based on the close relation between inference depth and path length. It is defined as the ratio of the number of inference steps to the number of entities involved. The higher this ratio, the more the reasoning process intuitively deviates from path-based reasoning. \cite{nora} found that all tested models struggled to generalize to reasoning problems with higher OPEC or BL than the training examples.

\paragraph{Learning to reason}
Let $\mathcal{P}$ be a set of Datalog rules. An example of a $\mathcal{P}$-reasoning problem is a tuple $(\mathcal{G},s,t,\mathcal{L})$ where $\mathcal{G}$ is a KG, $s$ and $t$ are entities from $\mathcal{G}$ (called the source and target entities), and $\mathcal{L}$ is the set of all relationships between $e$ and $f$ that can be inferred from $\mathcal{G}$, i.e.\ $\mathcal{L} = \{r \,|\, \mathcal{P}\cup\mathcal{G}\models r(e,f)\}$.
Given a set of training examples $(\mathcal{G}_1,s_1,t_1,\mathcal{L}_1),...,(\mathcal{G}_n,s_n,t_n,\mathcal{L}_n)$, the task of \emph{learning to reason} is to train a model which can answer queries of the form $(\mathcal{G}_{n+1},h_{n+1},t_{n+1},?)$, i.e.\ predict the relationships that hold between two entities in an unseen KG $\mathcal{G}_{n+1}$. All KGs $\mathcal{G}_i$ share the same set of relations $\mathcal{R}$, but their entities are distinct. Intuitively, the model thus has to learn the rules from $\mathcal{P}$ based on the given training examples, and then apply these rules to the unseen test examples. We typically evaluate models on test examples which are harder than the training examples, according to some metric of difficulty. This means that the models are required to reason about the learned rules, rather than relying on statistical shortcuts, and tests their ability to generalize in a systematic way.

In this paper, we will use Edge Transformers (ETs) for learning to reason \citep{edge-transformer}. While various competitive models have been proposed that can generalize well on path-based reasoning tasks \cite{DBLP:conf/icml/Minervini0SGR20,r5,cheng2023neural,khalid2025systematic}, most of these models struggle with off-path reasoning. ETs are theoretically capable of capturing arbitrary formulas from the logic $C_3$, i.e.\ the three-variable fragment of first-order logic with counting quantifiers \citep{DBLP:conf/nips/MullerKB024}, and can thus in principle learn the required knowledge. Empirically, ETs have also been found to generalize well in several settings \cite{edge-transformer,DBLP:conf/nips/MullerKB024}, and they achieved by far the strongest results on the NoRA benchmark \citep{nora}. 

%******************************************************
\section{Methodology}\label{secMethodology}

\paragraph{Problem setting}
Let $\mathsf{M}$ be a neural reasoning model that was trained on $\mathcal{P}$-reasoning examples (e.g.\ an ET). We want to find a sampling method which generates $\mathcal{P}$-reasoning examples that are hard for $\mathsf{M}$. Our main focus is on generating graphs $\mathcal{G}$ from which difficult queries can be constructed. To make this more precise, we want to find a sampler $\mathsf{S}$, i.e.\ a generative model from which we can sample worlds $\mathcal{G}$, satisfying a given set of constraint $\mathcal{C}$, which maximizes the following score:
\begin{align}\label{eqDifficulty}
\textit{difficulty}(\mathsf{S};\mathsf{M})= 1-\mathbb{E}_{\mathcal{G}\sim\mathsf{S}}[\min_{h,t \in \mathcal{E}_\mathcal{G}}\min_{r\in\mathcal{L}_{\mathcal{G},h,t}} \mathsf{M}_r(\mathcal{G},h,t)]
\end{align}
where we write $\mathcal{E}_{\mathcal{G}}$ for the set of entities in world $\mathcal{G}$, $\mathcal{L}_{\mathcal{G},h,t}$ for the ground truth answer to the query $(\mathcal{G},h,t,?)$, and  $\mathsf{M}_r(\mathcal{G},h,t)$ for the probability that $r(h,t)$ holds according to $\mathsf{M}$. 

\subsection{Evolutionary search}\label{secFunsearchMethodology}
We aim to learn samplers $\mathsf{S}$ that maximize $\textit{difficulty}(\mathsf{S};\mathsf{M})$ for a given model $\mathsf{M}$.
We focus on samplers that generate worlds by adding one edge at a time. To decide which edge to add in each step, we first randomly sample a set of $N_{\text{cand}}$ candidate triples $(e,r,f)$. These candidate triples are then scored by a \textbf{priority function} $\rho$, which depends on the partial world constructed thus far and the candidate triple. The highest scoring triple, according to this priority function, is added to the world. The problem of learning a sampler is thus reduced to that of learning a priority function.

To learn this priority function, we use an approach which is based on FunSearch \citep{funsearch}, an LLM-based evolutionary framework aimed at finding high-performing greedy algorithms for combinatorial problems. FunSearch is based on two key insights. First, given a candidate priority function, we can use an LLM to suggest an improved priority function (encoded as a Python method). Second, the quality of any given priority function $\rho$ can be robustly assessed using a task-specific evaluator. In our case, this evaluator uses the priority function $\rho$ to generate a number of KGs $\mathcal{G}_1,...,\mathcal{G}_{N_{\text{evalgraphs}}}$, and then uses the given model $\mathsf{M}$ to obtain an empirical estimate of \eqref{eqDifficulty}:
$$
\textit{fitness}(\rho)= 1 - \frac{1}{N_{\text{evalgraphs}}}\sum_{i=1}^{N_{\text{evalgraphs}}} \min_{h,t \in \mathcal{E}}\min_{r\in\mathcal{L}_{\mathcal{G}_i,h,t}} \mathsf{M}_r(\mathcal{G}_i,h,t)
$$
We refer to this estimate as the fitness of the priority function $\rho$. To obtain a priority function with a high fitness score, following FunSearch, we maintain a database of previously generated priority functions. In each step, we select two of these priority functions and ask the LLM to suggest an improved one, where the selection is biased towards the functions with the highest fitness. To encourage sufficient exploration, FunSearch maintains multiple databases of priority functions, called islands, where the evolution process is applied to each of these islands independently. After $N_{\text{iter}}$ iterations, half of the islands are removed, namely those whose best-scoring priority functions have the lowest fitness. These islands are then replaced by new databases, which are obtained by sampling the best-performing priority functions from the remaining islands. Each set of $N_{\text{iter}}$ iterations is referred to as a \textbf{cycle}. Further details can be found in Appendix \ref{appImprovingPriorityFunctionsDetails}.

The evolutionary approach is aimed at finding queries that are hard for some model $\mathsf{M}$. If this is successful, then a stronger model might be obtained by adding such queries to the training data. We want to analyze whether this process can be repeated. After a fixed number of $N_{\text{cycles}}$ have been completed, we therefore re-train the ET on a training set that includes queries that were sampled from the best overall priority function found so far. We then repeat the process with this new model. 
% We refer to each iteration of this process as a \textbf{round}.
We define a \textbf{round} as a period of $N_{\text{cycles}}$ during which a fixed model $\mathsf{M}$ is used, after which it is retrained.

%*********************
\subsection{Auto-research agents}\label{secAutoMethodology}
As a natural extension of the evolutionary approach, we consider an agent-in-the-loop procedure similar to \texttt{auto-research} \citep{autoresearch}. Different from the approach from Section \ref{secFunsearchMethodology}, the agent is able to build context over multiple experimental runs. Moreover, it has the ability to run code and perform rudimentary analysis via tool calls and inspections. We initiate the loop by telling the agent to read a \texttt{program.md} file, which is provided Appendix \ref{appAutoresearchPrompt}. For this experiment, we refactored our codebase to its minimal set (a simple ET training loop, a problem instance generator utilizing the sampling function, etc.) to  make sure the agent's context is minimally obfuscated. A coding agent (Claude Opus 4.6) iteratively edits a single Python program containing the priority function for a single round and the rest of protocol is kept the same as before. The agent's objective is to drive ET evaluation accuracy down while keeping the priority function as simple as possible.

%*********************
\subsection{Model-agnostic LLM-generated strategies}\label{secModelAgnosticSamplers}
We also wanted to assess the potential of directly relying on the coding abilities of frontier LLMs. To this end, we generated a number of different samplers through a dialogue with Claude Opus 4.6. 
% These samplers were obtained through a dialogue, where we first asked the model to generate a set of world rules, and then asked it to generate a sampler for these world rules. Subsequently, we asked to generalize the sampler to make it suitable for an arbitrary set of world rules $\mathcal{P}$ and constraints $\mathcal{C}$. We finally asked the model to improve on previously generated samplers and to come up with alternative strategies. 
This led to four general samplers: a sampler which uses hill-climbing to optimize graphs for deep derivations (\textbf{Cl-hill}), a sampler which uses backward chaining to ensure there is at least one deep derivation (\textbf{Cl-bw}), a sampler which builds graphs from small self-contained patterns which trigger specific derivation rules (\textbf{Cl-motif}), and a hybrid strategy obtained by prompting Claude to combine ideas from the three aforementioned methods with the specific aim to challenge GNNs (\textbf{Cl-atlas}). These samplers can, in principle, be used for any set of Datalog rules $\mathcal{P}$ and constraints $\mathcal{C}$. We also obtained, as part of this dialogue, a number of samplers that were specific to the NoRA world: a sampler which starts from a few family skeleton templates (\textbf{Cl-N-temp}), a sampler which starts from a minimal two-person married seed and greedily expands the graph one edge at a time (\textbf{Cl-N-greedy}), and a variant of Cl-bw with heuristics that are specific to the NoRA world (\textbf{Cl-N-bw}). Further details on these samplers, and how they were obtained, can be found in Appendix \ref{appGeneratingSamplersDetails}.
%******************************************************
\section{Experiments}

% \subsection{Experimental setup}

% Focus on small but hard problem instances! 

Reasoning models can be straightforwardly challenged by presenting them with sufficiently large KGs. Our main focus is therefore on whether it is possible to find \emph{small} KGs which challenge these models. In particular, we restrict our experiments to KGs of at most 8 entities.

Our analysis primarily focuses on the rules and constraints from the NoRA 1.1 benchmark \cite{nora}. Similar to CLUTRR, this benchmark focuses on the problem of reasoning about family relationships. However, in contrast to previous benchmarks, NoRA requires off-path reasoning, which poses particular challenges for existing methods. NoRA 1.1 consists of a training split and three test splits: test-D, test-OPEC and test-BL (consisting of examples that are systematically harder than the training examples in terms of one of the difficulty metrics). For our analysis, some of the examples from test-D, test-OPEC and test-BL will be used for training, since we want to find problems with different forms of difficulty than captured by these test sets. 

To allow us to validate our findings on a second domain, we used Claude to generate a completely new set of world rules and constraints (see Appendix \ref{appGeneratingWorldRulesDetails}). The resulting domain is called \textbf{Iron Coast}. In addition to providing another dataset for evaluating the considered sampling strategies, experiments on this domain will allow us to assess the usefulness of LLM-generated reasoning domains, and thus whether it is feasible to fully automate the process of evaluating neural relational reasoners. 

Results are evaluated in terms of exact-match accuracy, i.e.\ we report the fraction of queries $(\mathcal{G},h,t,?)$ for which the predicted set of relations exactly matches the ground truth $\mathcal{L}_{\mathcal{G},h,t}$.

\begin{table}[t]
\centering
\setlength\tabcolsep{4pt}
\footnotesize
\caption{Cross-evaluation of samplers on NoRA in terms of exact-match accuracy, comparing Claude-generated samplers with the result of four independent runs of the evolutionary process with qwen3-coder-next-80b.}\label{tabCrossEvaluationSamplers}
\begin{tabular}{ll ccccccccc} 
\toprule
&& \multicolumn{9}{c}{\textbf{Test set}}\\
\cmidrule(lr){3-11}
&& \textbf{Cl-atlas} & \textbf{Cl-motif} & \textbf{Cl-N-bw} & \textbf{Cl-N-greedy} & \textbf{Cl-N-temp} & \textbf{Evo1} & \textbf{Evo2} & \textbf{Evo3} & \textbf{Evo4} \\
\midrule
\multirow{9}{*}{\rotatebox[origin=c]{90}{\textbf{Training set}} } & \textbf{Cl-atlas} & \heat{0.92} & \heat{0.59} & \heat{0.26} & \heat{0.02} & \heat{0.28} & \heat{0.16} & \heat{0.17} & \heat{0.16} & \heat{0.16} \\
& \textbf{Cl-motif} & \heat{0.58} & \heat{0.96} & \heat{0.43} & \heat{0.19} & \heat{0.32} & \heat{0.24} & \heat{0.25} & \heat{0.25} & \heat{0.24} \\
& \textbf{Cl-N-bw} & \heat{0.04} & \heat{0.09} & \heat{0.99} & \heat{0.10} & \heat{0.13} & \heat{0.08} & \heat{0.09} & \heat{0.06} & \heat{0.09} \\
& \textbf{Cl-N-greedy} & \heat{0.05} & \heat{0.10} & \heat{0.34} & \heat{0.96} & \heat{0.16} & \heat{0.05} & \heat{0.06} & \heat{0.07} & \heat{0.05} \\
& \textbf{Cl-N-temp} & \heat{0.01} & \heat{0.03} & \heat{0.06} & \heat{0.26} & \heat{0.98} & \heat{0.01} & \heat{0.02} & \heat{0.02} & \heat{0.02} \\
& \textbf{Evo1} & \heat{0.18} & \heat{0.23} & \heat{0.23} & \heat{0.18} & \heat{0.29} & \heat{0.68} & \heat{0.50} & \heat{0.52} & \heat{0.50} \\
& \textbf{Evo2} & \heat{0.17} & \heat{0.23} & \heat{0.28} & \heat{0.19} & \heat{0.20} & \heat{0.53} & \heat{0.62} & \heat{0.54} & \heat{0.60} \\
& \textbf{Evo3} & \heat{0.17} & \heat{0.23} & \heat{0.31} & \heat{0.19} & \heat{0.31} & \heat{0.52} & \heat{0.53} & \heat{0.55} & \heat{0.54} \\
& \textbf{Evo4} & \heat{0.17} & \heat{0.23} & \heat{0.15} & \heat{0.15} & \heat{0.36} & \heat{0.52} & \heat{0.55} & \heat{0.55} & \heat{0.52} \\
\bottomrule
\end{tabular}
\end{table}

\begin{figure}[t]    
    \centering
    \includegraphics[width=\linewidth]{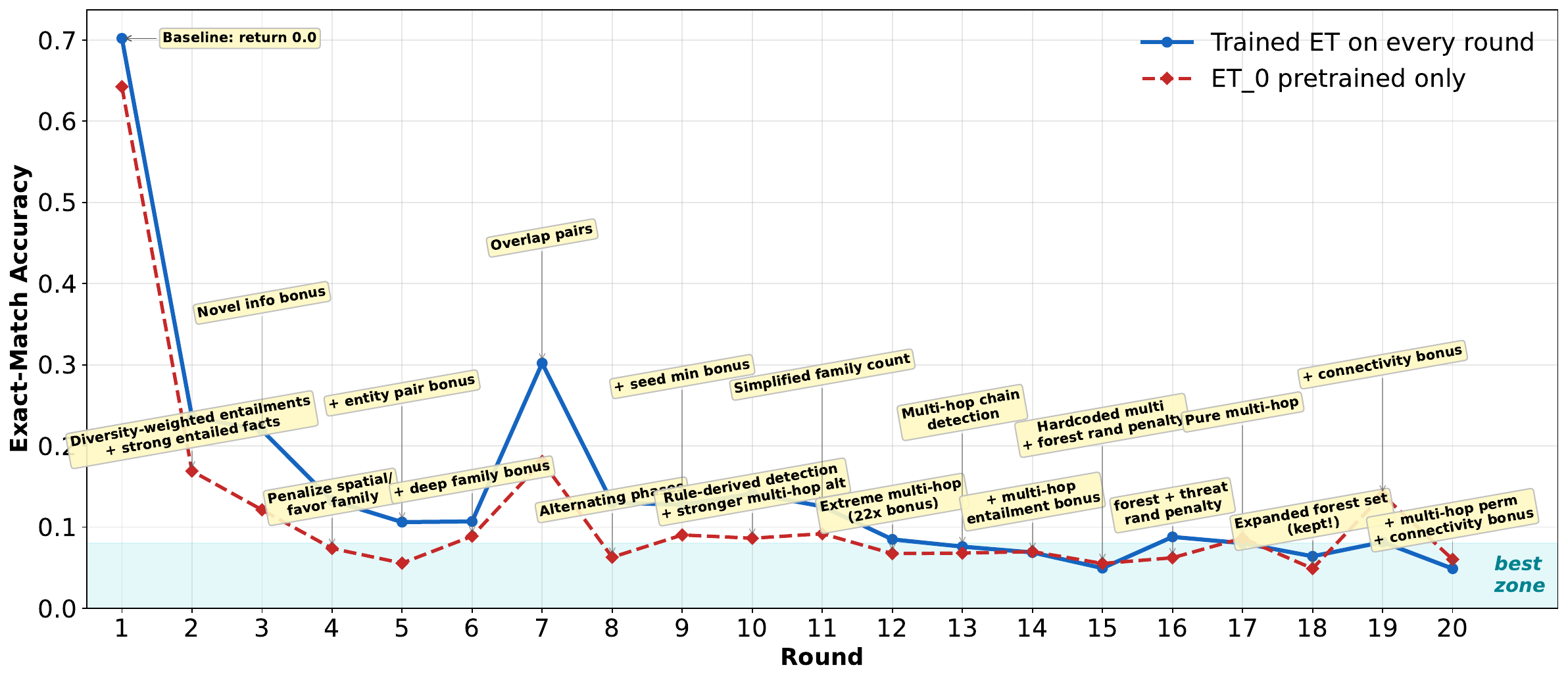}
    \caption{Auto-research results for Claude Opus 4.6 coding agent running in an infinite agent-in-the-loop. Exact-Match accuracies of the ET$_0$ (trained only once on NoRA 1.1) and the ET that is trained continually on each successive rounds from fresh and historical data generated by each round's priority function can be seen. Summaries of major changes that are introduced per round are also annotated. }
    \label{fig:auto-research}
\end{figure}

\paragraph{Evaluation of sampling strategies}

The difficulty of the queries that are obtained using a given sampling strategy can only be determined relative to a particular model. In Table \ref{tabCrossEvaluationSamplers}, we therefore present a \emph{cross-evaluation} of different samplers. Each entry in the table shows the performance of an ET that was trained on examples from a given training set and evaluated on queries generated by a particular sampler. 
We have included 5 of the Claude-generated samplers \footnote{The remaining two generic Claude samplers generated non-nonsensical worlds, which is related to the fact that the NoRA constraints do not exhaustively avoid all nonsensical worlds. They were thus omitted for this analysis.}, which were used both to generate training sets and test sets. For instance, the results in the row marked Cl-atlas are those for an ET that was \emph{trained} on examples sampled from Cl-atlas. The column marked Cl-atlas shows the results that were obtained for \emph{test queries} that were sampled from Cl-atlas. We also included samplers obtained from four independent runs of the  evolutionary process with Qwen3-next-80b. In particular, the row marked Evo$i$ show results for an ET that was trained on the round 4 training set of a evolutionary process with initial seed $i$. The columns marked Evo$i$ show results for test queries that were sampled using the priority function developed in round 5. Note that the diagonal elements are thus in-distribution for the Claude samplers, but not for the evolutionary runs.
Full details on the experimental set-up can be found in Appendix \ref{appDetailsExperimentalMethodology}.

The results show several clear patterns. First, the samplers obtained from the four runs of the evolutionary process appear to be similar: models trained on Evo1--Evo4 achieve a similar performance on the Evo1--Evo4 test sets. In contrast, models trained on the evolutionary samplers perform poorly on the Claude datasets, and \emph{vice versa}. Among the Claude-based samplers, training on \emph{Cl-motif} examples overall achieves the best results, whereas the NoRA-specific samplers are the weakest in this respect. As expected, the Claude models perform nearly perfectly on in-distribution examples.

We performed a similar cross-evaluation of samplers on the Iron Coast domain, finding broadly the same patterns (see Appendix \ref{secEvaluationIronCoast}). Furthermore, changing the LLM from Qwen3-next-80b to gpt-oss-120b or deepseek-coder-33b only had a minimal impact (see Appendix \ref{secEvaluationLLMComparison}).

Figure \ref{fig:auto-research} analyzes the effectiveness of the Auto-research experiment, described in Section \ref{secAutoMethodology}. In each round of this process, a priority function is obtained. The figure shows the performance of two ETs on queries sampled from that priority function: the initial ET (trained on the NoRA dataset, as for the evolutionary approach) and an ET which is trained on a combination of the initial training data and examples sampled from the priority functions from previous rounds. As can be seen, most of the adversarial gain occurs in the first 3 rounds. Towards the later rounds, the performance of both ETs converges, suggesting that the Auto-research agent has found a priority function that challenges the ET, even when recent variations of that priority function are used for training the model.

\begin{figure}
\begin{subfigure}[b]{0.37\textwidth}
\includegraphics[width=\linewidth]{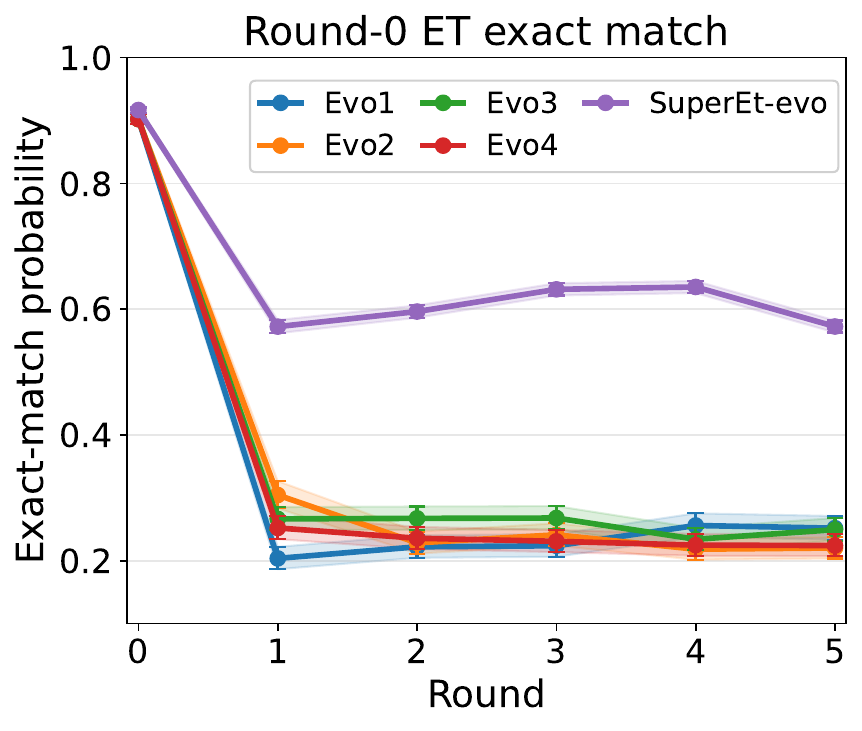}
\caption{\label{figEvolutionSuperET}}
\end{subfigure}
\hfill
\begin{subfigure}[b]{0.63\textwidth}
\includegraphics[width=\linewidth]{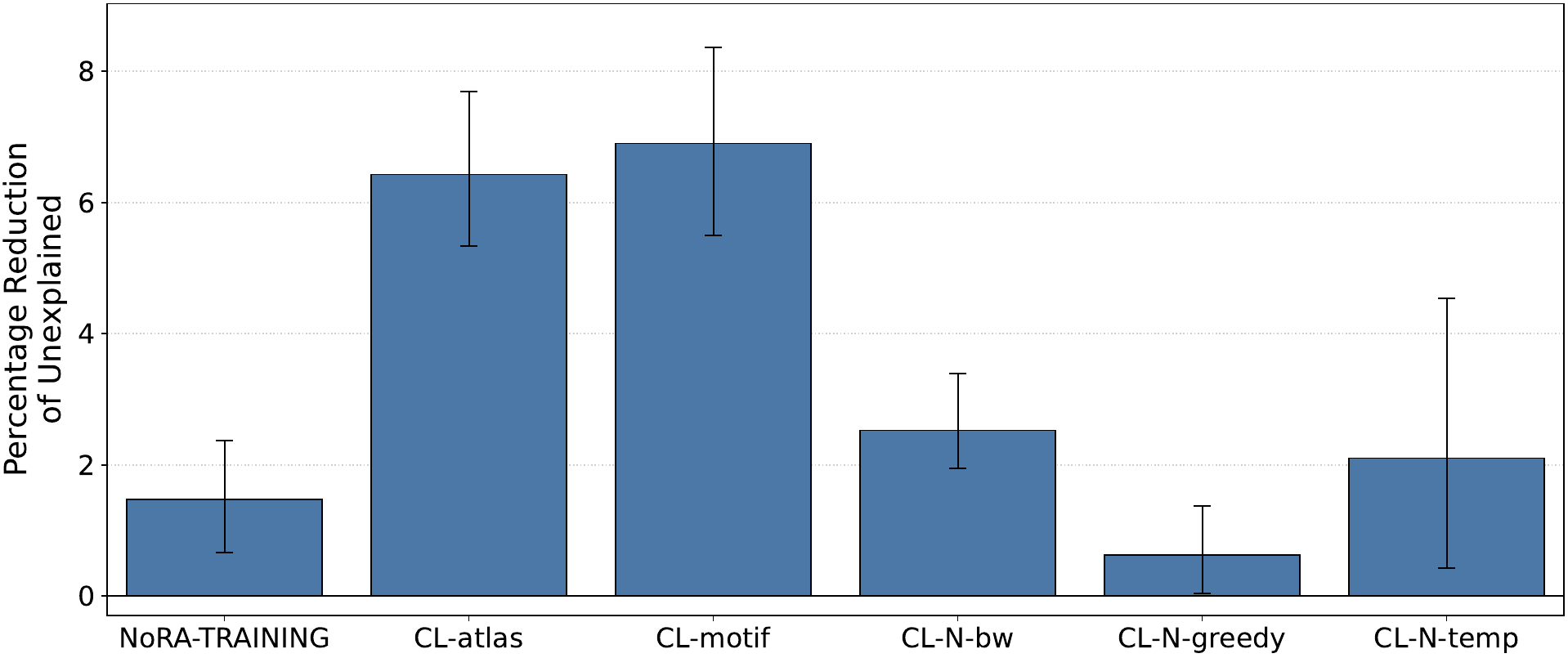}
\vspace{1pt}
\caption{\label{figDifficultyAnalysisEntailedOPEC}}
\end{subfigure}
\caption{(a) Performance of the original ET after each round of the evolutionary process, comparing a variant initialized with the \textsc{SuperET} with four runs that were initialized with the standard ET. The results reflect the performance of the initial model on samples generated by the best priority functions at each round, and are reported in terms of exact-match accuracy. (b) Analysis of percentage reduction in unexplained difficulty when incorporating entailed off-path edges (using queries sampled with priority functions obtained at the end of various standard evolutionary runs.).}
\end{figure}

\begin{table}
\footnotesize
\centering
\caption{Cross-evaluation of priority functions obtained through evolutionary search (for NoRA) using qwen3-coder-next-80b, the priority function obtained at round 20 of the auto-research agent, and the priority function obtained when the process was initialised with the \textsc{SuperET} model.  Results are reported in terms of exact-match accuracy.}\label{tabCrossEvaluationSuperET}

\begin{tabular}{llcccccc}
\toprule
&& \multicolumn{6}{c}{\textbf{Test set}}\\
\cmidrule(lr){3-8}
 && \textbf{Evo1} & \textbf{Evo2} & \textbf{Evo3} & \textbf{Evo4} & \textbf{Auto-research} & \textbf{\textsc{SuperET}-evo} \\
\midrule
\multirow{6}{*}{\rotatebox[origin=c]{90}{\textbf{Training set}} } &\textbf{Evo1} & \heat{0.68} & \heat{0.50} & \heat{0.52} & \heat{0.50} & \heat{0.03} & \heat{0.47} \\
&\textbf{Evo2} & \heat{0.53} & \heat{0.62} & \heat{0.54} & \heat{0.60} & \heat{0.03} & \heat{0.47} \\
&\textbf{Evo3} & \heat{0.52} & \heat{0.53} & \heat{0.55} & \heat{0.54} & \heat{0.04} & \heat{0.45} \\
&\textbf{Evo4} & \heat{0.52} & \heat{0.55} & \heat{0.55} & \heat{0.52} & \heat{0.03} & \heat{0.46} \\
&\textbf{Auto-research} & \heat{0.12} & \heat{0.09} & \heat{0.13} & \heat{0.09} & \heat{0.15} & \heat{0.11} \\
&\textbf{\textsc{SuperET}-evo} & \heat{0.86} & \heat{0.86} & \heat{0.87} & \heat{0.87} & \heat{0.57} & \heat{0.70} \\
\bottomrule
\end{tabular}
\end{table}

%****************
\paragraph{Training reasoning models on augmented training data}
In Table \ref{tabCrossEvaluationSamplers}, we found that both the evolved priority functions and the model-agnostic LLM-generated samplers succeed in challenging the ET. We now investigate whether a more robust ET model can be obtained by combining training examples from all these samplers. %In other words, we want to analyze whether the failures of the ET are due to limitations in the training data, or whether they suggest more fundamental limitations. 
Specifically, for this experiment we train a model from 13 sources: the 9 training sets considered in Table \ref{tabCrossEvaluationSamplers}, together with the training set, test-OPEC, test-BL and test-D splits from NoRA 1.1. We sampled 1300 examples at random from each of these sources (which reflects the size of the smallest source). In the following, we refer to the resulting model as \textsc{SuperET}. We then repeated the evolutionary process for this \textsc{SuperET}. 
Figure \ref{figEvolutionSuperET} shows the performance of the (initial) \textsc{SuperET} after each round of the evolutionary process (in purple), together with the performance of the (initial) ET in four runs of the standard evolutionary process. As can be seen, the performance of the \textsc{SuperET} remains relatively strong throughout the process, showing that the evolutionary approach struggles to find strategies that challenge this model. 

Table \ref{tabCrossEvaluationSuperET} presents a cross-evaluation of 6 models: ETs trained on the round-4 training data from four standard runs of the evolutionary process (which are distinct from those that were used for training the \textsc{SuperET}), a model trained on the round-19 training set from the Auto-research experiment, and the round-4 training data from the evolutionary process with the \textsc{SuperET}. The models are evaluated on the round-5 priority functions of the evolutionary approaches and the round-20 priority function from the Auto-research experiment.

This analysis confirms the robustness of \textsc{SuperET}-evo, i.e.\ the \textsc{SuperET} variant obtained after run 4 of the evolutionary process. We find that, apart from \textsc{SuperET}-evo, all models perform poorly on the queries obtained from the Auto-research sampler. We see that the model trained on queries from the Auto-research sampler performs poorly overall, suggesting a lack of diversity in the generated worlds.

\begin{table}
\centering
\footnotesize
\caption{Analysis of unexplained difficulty, in terms of the score defined in \eqref{eqUnexplained}. \label{tabDifficultyAnalysis}}
\begin{tabular}{llcccccc}
\toprule
&& \multicolumn{6}{c}{\textbf{Queries}}\\
\cmidrule(lr){3-8}
&& \textbf{Evo} & \textbf{Claude} & \textbf{Auto-research} & \textbf{\textsc{SuperET}-evo} & \textbf{test-OPEC} & \textbf{test-D} \\
\midrule
\multirow{5}{*}{\rotatebox[origin=c]{90}{\textbf{Model}} } & \textbf{Evo}             & \heat{0.84} & \heat{0.79} & \heat{0.88} & \heat{0.80} & \heat{0.59} & \heat{0.60}\\
& \textbf{Cl-atlas}        & \heat{0.54} & \heat{0.81} & \heat{0.87} & \heat{0.53} & \heat{0.45} & \heat{0.44}\\
&\textbf{Auto-research}   & \heat{0.64} & \heat{0.79} & \heat{0.86} & \heat{0.70} & \heat{0.53} & \heat{0.54}\\
& \textbf{\textsc{SuperET}-evo}     & \heat{0.90} & \heat{0.79} & \heat{0.79} & \heat{0.89} & \heat{0.61} & \heat{0.61}\\
& \textbf{NoRA-training}   & \heat{0.59} & \heat{0.77} & \heat{0.76} & \heat{0.66} & \heat{0.51} & \heat{0.49}\\
\bottomrule
\end{tabular}
\end{table}

%****************
\paragraph{Analysis}
We analyze to what extent the considered samplers have identified novel forms of difficulty, beyond inference depth, OPEC and BL. For a given model and test set, we train a classifier to predict whether the model will answer a given query correctly or not, using as features only the inference depth, OPEC and BL of that query (see Appendix \ref{secDetailsProblemDifficulty} for details). We then evaluate the extent to which problem difficulty is unexplained, based on the McFadden pseudo-R\textsuperscript{2} ($R^2_{\text{McF}}$), as follows:
\begin{align}\label{eqUnexplained}
\textit{Unexplained} = 1- R^2_{\text{McF}} = \frac{l_{\textit{model}}}{l_{\textit{null}}} = \frac{\sum_{q\in Q^+} \log p_{\textit{model}}(q) + \sum_{q\in Q^-} \log (1-p_{\textit{model}}(q))}{\sum_{q\in Q^+} \log p_{\textit{null}}(q) + \sum_{q\in Q^-} \log (1-p_{\textit{null}}(q))}
\end{align}
where $Q^+$ and $Q^-$ are the queries that were answered correctly and incorrectly by the ET, for a given set of queries obtained from the considered sampler; $p_{\textit{model}}(q)$ is the probability assigned by our classifier that $q$ will be answered correctly; and $p_{\textit{null}}(q)$  is the probability assigned by an uninformed model, i.e.\ $p_{\textit{null}}(q)= \frac{|Q^+|}{|Q^+| + |Q^-|}$. In Table \ref{tabDifficultyAnalysis}, we report the  \textit{Unexplained} value for different models and sampling strategies. This value reflects the extent to which there are aspects of difficulty in the queries that are unexplained by inference depth, OPEC and BL. A value of 1 reflects a situation where knowing the inference depth, OPEC and BL of a query does not provide any information about its difficulty. In contrast, a value of 0 reflects a situation where these difficulty metrics fully determine whether a query is answered correctly. 
Each row in the table corresponds to an ET that was trained on a particular training set: the round-5 training data of a standard evolutionary run, queries sampled from Cl-Atlas, queries sampled using the round-20 training data from the auto-research experiment, the round-5 training data of an evolutionary run based on the \textsc{SuperET}, and the standard NoRA 1.1 training split. The columns correspond to different samplers that were used for generating the test queries: the round-5 priority function from a standard evolutionary run, Cl-Atlas, the round-20 sampler of the auto-research experiment, the round-5 priority function from the evolutionary process based on the \textsc{SuperET} and the test-OPEC and test-D test splits from NoRA 1.1.

As expected, the \emph{Unexplained} values for the test-OPEC and test-D queries are among the lowest in the table. In contrast, the values for the \emph{Claude} and \emph{Auto-research} test sets are consistently high, for all models, showing that these samplers generate problems whose difficulty is poorly explained by the considered metrics. Finally, for Evo and \textsc{SuperET}-evo, we see large discrepancies across the different models. Focussing on the \textsc{SuperET}-evo model, for instance, we can see a high unexplained value for the Evo queries. Given the stronger overall performance of this model, this suggests that it performs better than the other models on problems that are only difficult because of their inference depth, OPEC or BL value. The problems on which it struggles, however, have a difficulty which cannot be explained by these metrics.

When inspecting the generated problem instances from the evolutionary samplers, we identified a class of queries that are difficult because the derivation relies on inferred off-path edges. Figure \ref{fig1a} shows an example of such a query, where the \textit{is\_female} loop is an off-path edge.  However, since this edge is inferred, the OPEC value of this query is 0. Despite only involving 5 entities, all models included in the analysis from Figure \ref{figDifficultyAnalysisEntailedOPEC} misclassified this example. More generally, we found that incorporating the number of such inferred off-path edges can reduce the value of the \textit{Unexplained} score. Specifically, let us defined the \textbf{Off-Path Entailed Edge Count (OPEEC)} as the number of entailed off-path edges that are used to derive the answer. Figure \ref{figDifficultyAnalysisEntailedOPEC} shows the percentage reduction in the \textit{Unexplained} score when adding OPEEC as a feature, for a number of different models evaluated on  test queries obtained by combining, for each evolutionary run, the queries sampled using the priority function produced at the end of round 5. (see Appendix \ref{secDetailsProblemDifficulty} for further detail).

%******************************************************
\section{Related Work}

\paragraph{Scientific discovery and meta-learning}
LLMs for automating scientific discovery via methods like FunSearch \citep{funsearch} and its successor AlphaEvolve \citep{DBLP:journals/corr/abs-2506-13131} (essentially its multi-objective and multi-file editing counterpart) have yielded promising results in various domains including mathematics \citep{funsearch-math}, Bayesian and Graph-based Optimization \citep{funbo, grail-funsearch} and cognitive science \citep{funsearch-cognitive}. They focus on problems with automatically verifiable solutions, e.g., search for mathematical objects with certain properties or designing heuristic algorithms. 
At its heart, the idea is much older and is related to meta-learning \citep{schmidhuber1987evolutionary}, which attempts to learn algorithms via end-to-end optimization but is encumbered, primarily, by the intrinsic inability to differentiate through complex learning processes such as reinforcement learning (RL). Here, evolution-based optimization is a scalable alternative to RL \citep{salimans2017evolution} due its gradient-free nature and has, in fact, also been applied to evolve surrogate objective functions to discover novel loss functions in RL \citep{discovering-PO, jackson2024discovering, discover-loss-lu-2}. Within a wider setting, LLM-based evolution has been applied to prompt engineering \citep{guo2024connecting,agrawal2026gepa}, neural architecture search \citep{DBLP:conf/gecco/NasirETJC24} and reward shaping \citep{DBLP:conf/iclr/HazraSPLM25, evil}.

\paragraph{Self-improvement and adversarial learning}
The idea of self-improvement \citep{schmidhuber1987evolutionary, nivel2013bounded} is closely related to self-play \citep{samuel1959some, sutton1998reinforcement}, which naturally admits an adversarial interpretation \citep{gans, arjovsky2017wasserstein} in the form of a two-player minimax game. Our approach is similar in spirit to adversarial learning methods, particularly adversarial imitation learning \citep{gans, ail-ghasemipour20a, discovering-adversarial} where reward assignment functions are shaped as a function of the outputs of the underlying policy neural network whose training stability and performance are optimized. Another related thread is the idea of self-challenging \citep{zhou2026selfchallenging, kulikov2026autodata} or self-play \citep{sukhbaatar2018intrinsic} where a challenger generates training examples for a solver, which can both be optimized together, thereby iteratively producing harder training examples. \cite{nie-etal-2020-adversarial} have also created a benchmark that specifically challenges a given model, focusing on the task of Natural Language Inference (NLI) and relying on crowdworkers to obtain the problem. 

\paragraph{Synthetic benchmark generation}
The use of LLMs for generating synthetic benchmarks is common practice, e.g., instruction-tuning data for QA and agentic tool-use \citep{schick-schutze-2021-generating, self-instruct, shi2026taskcraft, agentgym} due to inherent scalability of the approach and the scarcity of human data \citep{singh2024beyond}. For instance, \cite{DBLP:conf/icml/GuinetODC24} generate question and answer pairs for evaluating Retrieval Augmented Generation (RAG) pipelines. 
The use of synthetic datasets is also common practice for evaluating the reasoning abilities of LLMs in various settings \citep{DBLP:conf/nips/DziriLSLJLWWB0H23,valmeekam2022large,DBLP:conf/iclr/MirzadehASTBF25,DBLP:journals/corr/abs-2506-06941,DBLP:conf/ijcnlp/XieHZYCLLGK25}. Seedless, self-challenging, agentic data synthesis using a top-down, taxonomic, self-verification loop have also been proposed \citep{davidson2026reasoningdriven, kulikov2026autodata}.
However, the evaluation pipelines in such settings rely on predefined sampling strategies, focussing more on generating test examples that cannot have appeared in pre-training than on discovering particular dimensions of difficulty. Another concern is verifying the quality of LLM-based problem instances. In contrast, our approach uses LLM-sampled programs that generate self-challenging instances.

%******************************************************
\section{Conclusions}
We have studied the potential of LLMs for learning to sample problem instances that are challenging for relational reasoning models. Our analysis has focused on three broad strategies: (i) LLM-driven evolutionary search based on FunSearch, (ii) Auto-research agents which autonomously evolve sampling strategies, and (iii) simple dialogue-based prompting of frontier LLMs. We found all strategies to be successful in challenging the Edge Transformer (ET), a state-of-the-art neural relational learning model, despite the fact that the generated problem instances are small (consisting of at most 8 entities). We then considered whether a more robust ET model could be obtained by incorporating the learned samplers into the training data. The resulting model (\textsc{SuperET}) was indeed significantly more robust against the FunSearch-based evolutionary process, although it was still challenged by the sampler that was learned by the Auto-research agent. Our analysis of the learned samplers revealed that they have (implicitly) discovered aspects of difficulty which go beyond what can be explained through known difficulty metrics. 
Overall, our results show that the considered sampling strategies can play a central role in research on neural relational reasoning, both for stress-testing the robustness of trained models and for generating diverse and informative training examples. In the considered setting, the world rules are assumed to be given, i.e.\ we use LLM-driven strategies to sample knowledge graphs, but not to sample the rules that determine the regularities that need to be learned. However, we found that this approach can also be successfully applied to LLM-generated world rules. This makes it possible to fully automate the evaluation loop.

\paragraph{Limitations}
Our analysis leaves a number of important questions for future work. For instance, we may wonder to what extent the learned priority functions transfer to other sets of world rules. In other words, are the learned priority functions challenging because they exploit particular peculiarities of the considered NoRA world, or do they capture deeper insights into what makes reasoning problems challenging for ETs. Furthermore, we have limited our analysis to sets of Datalog rules, which means that challenging forms of reasoning, such as non-monotonic reasoning and reasoning with disjunctive rules, are not covered. Our preliminary analysis showed that LLMs can generate interesting sets of world rules that require such forms of reasoning (using constructs such as negation-as-failure and choice rules), suggesting that our approach can indeed be extended along these lines. While we believe the ET is a natural choice for the considered setting, our analysis could be extended to a number of alternative reasoning models.

% \begin{ack}
% Use unnumbered first level headings for the acknowledgments. All acknowledgments

% Do {\bf not} include this section in the anonymized submission, only in the final paper. You can use the \texttt{ack} environment provided in the style file to automatically hide this section in the anonymized submission.
% \end{ack}
\newpage
\bibliographystyle{plainnat}
\bibliography{references}

@article{funbo,
  title={Funbo: Discovering acquisition functions for bayesian optimization with funsearch},
  author={Aglietti, Virginia and Ktena, Ira and Schrouff, Jessica and Sgouritsa, Eleni and Ruiz, Francisco JR and Malek, Alan and Bellot, Alexis and Chiappa, Silvia},
  journal={arXiv preprint arXiv:2406.04824},
  year={2024}
}

@inproceedings{
discovering-adversarial,
title={On Discovering Algorithms for Adversarial Imitation Learning},
author={Shashank Reddy Chirra and Jayden Teoh and Praveen Paruchuri and Pradeep Varakantham},
booktitle={The Fourteenth International Conference on Learning Representations},
year={2026},
url={https://openreview.net/forum?id=hPz3doftL4}
}

@InProceedings{funsearch-cognitive,
  title = 	 {Discovering Symbolic Cognitive Models from Human and Animal Behavior},
  author =       {Castro, Pablo Samuel and Tomasev, Nenad and Anand, Ankit and Sharma, Navodita and Mohanta, Rishika and Dev, Aparna and Perlin, Kuba and Jain, Siddhant and Levin, Kyle and Elteto, Noemi and Dabney, Will and Novikov, Alexander and Turner, Glenn C and Eckstein, Maria K and Daw, Nathaniel D. and Miller, Kevin J and Stachenfeld, Kim},
  booktitle = 	 {Proceedings of the 42nd International Conference on Machine Learning},
  pages = 	 {6849--6890},
  year = 	 {2025},
  editor = 	 {Singh, Aarti and Fazel, Maryam and Hsu, Daniel and Lacoste-Julien, Simon and Berkenkamp, Felix and Maharaj, Tegan and Wagstaff, Kiri and Zhu, Jerry},
  volume = 	 {267},
  series = 	 {Proceedings of Machine Learning Research},
  month = 	 {13--19 Jul},
  publisher =    {PMLR},
  url = 	 {https://proceedings.mlr.press/v267/castro25a.html},
}

@inproceedings{
grail-funsearch,
title={{GRAIL}: Graph Edit Distance and Node Alignment using {LLM}-Generated Code},
author={Samidha Verma and Arushi Goyal and Ananya Mathur and Ankit Anand and Sayan Ranu},
booktitle={Forty-second International Conference on Machine Learning},
year={2025},
url={https://openreview.net/forum?id=NzoZXju2bL}
}

@article{funsearch-math,
  title={Generative modeling for mathematical discovery},
  author={Ellenberg, Jordan S and Fraser-Taliente, Cristofero S and Harvey, Thomas R and Srivastava, Karan and Sutherland, Andrew V},
  journal={arXiv preprint arXiv:2503.11061},
  year={2025}
}

@phdthesis{schmidhuber1987evolutionary,
  title={Evolutionary principles in self-referential learning, or on learning how to learn: the meta-meta-... hook},
  author={Schmidhuber, J{\"u}rgen},
  year={1987},
  school={Technische Universit{\"a}t M{\"u}nchen}
}

@article{discovering-PO,
  title={Discovered policy optimisation},
  author={Lu, Chris and Kuba, Jakub and Letcher, Alistair and Metz, Luke and Schroeder de Witt, Christian and Foerster, Jakob},
  journal={Advances in Neural Information Processing Systems},
  volume={35},
  pages={16455--16468},
  year={2022}
}

@inproceedings{
jackson2024discovering,
title={Discovering Temporally-Aware Reinforcement Learning Algorithms},
author={Matthew Thomas Jackson and Chris Lu and Louis Kirsch and Robert Tjarko Lange and Shimon Whiteson and Jakob Nicolaus Foerster},
booktitle={The Twelfth International Conference on Learning Representations},
year={2024},
url={https://openreview.net/forum?id=MJJcs3zbmi}
}

@article{salimans2017evolution,
  title={Evolution strategies as a scalable alternative to reinforcement learning},
  author={Salimans, Tim and Ho, Jonathan and Chen, Xi and Sidor, Szymon and Sutskever, Ilya},
  journal={arXiv preprint arXiv:1703.03864},
  year={2017}
}

@inproceedings{gans,
 author = {Goodfellow, Ian J. and Pouget-Abadie, Jean and Mirza, Mehdi and Xu, Bing and Warde-Farley, David and Ozair, Sherjil and Courville, Aaron and Bengio, Yoshua},
 booktitle = {Advances in Neural Information Processing Systems},
 editor = {Z. Ghahramani and M. Welling and C. Cortes and N. Lawrence and K. Weinberger},
 pages = {},
 publisher = {Curran Associates, Inc.},
 title = {Generative Adversarial Nets},
 url = {https://proceedings.neurips.cc/paper_files/paper/2014/file/f033ed80deb0234979a61f95710dbe25-Paper.pdf},
 volume = {27},
 year = {2014}
}

@InProceedings{evil,
  title = 	 {{E}v{IL}: Evolution Strategies for Generalisable Imitation Learning},
  author =       {Sapora, Silvia and Swamy, Gokul and Lu, Chris and Teh, Yee Whye and Foerster, Jakob Nicolaus},
  booktitle = 	 {Proceedings of the 41st International Conference on Machine Learning},
  pages = 	 {43407--43421},
  year = 	 {2024},
  editor = 	 {Salakhutdinov, Ruslan and Kolter, Zico and Heller, Katherine and Weller, Adrian and Oliver, Nuria and Scarlett, Jonathan and Berkenkamp, Felix},
  volume = 	 {235},
  series = 	 {Proceedings of Machine Learning Research},
  month = 	 {21--27 Jul},
  publisher =    {PMLR},
  url = 	 {https://proceedings.mlr.press/v235/sapora24a.html},
}

@inproceedings{self-instruct,
    title = "Self-Instruct: Aligning Language Models with Self-Generated Instructions",
    author = "Wang, Yizhong  and
      Kordi, Yeganeh  and
      Mishra, Swaroop  and
      Liu, Alisa  and
      Smith, Noah A.  and
      Khashabi, Daniel  and
      Hajishirzi, Hannaneh",
    editor = "Rogers, Anna  and
      Boyd-Graber, Jordan  and
      Okazaki, Naoaki",
    booktitle = "Proceedings of the 61st Annual Meeting of the Association for Computational Linguistics (Volume 1: Long Papers)",
    month = jul,
    year = "2023",
    address = "Toronto, Canada",
    publisher = "Association for Computational Linguistics",
    url = "https://aclanthology.org/2023.acl-long.754/",
    doi = "10.18653/v1/2023.acl-long.754",
    pages = "13484--13508",
}

@article{
davidson2026reasoningdriven,
title={Reasoning-Driven Synthetic Data Generation and Evaluation},
author={Tim R. Davidson and Benoit Seguin and Enrico Bacis and Cesar Ilharco and Hamza Harkous},
journal={Transactions on Machine Learning Research},
issn={2835-8856},
year={2026},
url={https://openreview.net/forum?id=NALsdGEPhB},
note={J2C Certification}
}

@inproceedings{
shi2026taskcraft,
title={TaskCraft: Automated Generation of Agentic Tasks},
author={Dingfeng Shi and Jingyi Cao and Qianben Chen and Weichen Sun and Weizhen Li and Hongxuan Lu and Fangchen Dong and Tianrui Qin and King Zhu and Minghao Liu and Yuchen Eleanor Jiang and Jian Yang and Ge Zhang and Jiaheng Liu and Changwang Zhang and Jun Wang and Wangchunshu Zhou},
booktitle={The Fourteenth International Conference on Learning Representations},
year={2026},
url={https://openreview.net/forum?id=UJFCyrYM1V}
}

@inproceedings{schick-schutze-2021-generating,
    title = "Generating Datasets with Pretrained Language Models",
    author = {Schick, Timo  and
      Sch{\"u}tze, Hinrich},
    editor = "Moens, Marie-Francine  and
      Huang, Xuanjing  and
      Specia, Lucia  and
      Yih, Scott Wen-tau",
    booktitle = "Proceedings of the 2021 Conference on Empirical Methods in Natural Language Processing",
    month = nov,
    year = "2021",
    address = "Online and Punta Cana, Dominican Republic",
    publisher = "Association for Computational Linguistics",
    url = "https://aclanthology.org/2021.emnlp-main.555/",
    doi = "10.18653/v1/2021.emnlp-main.555",
    pages = "6943--6951",
}

@article{
singh2024beyond,
title={Beyond Human Data: Scaling Self-Training for Problem-Solving with Language Models},
author={Avi Singh and John D Co-Reyes and Rishabh Agarwal and Ankesh Anand and Piyush Patil and Xavier Garcia and Peter J Liu and James Harrison and Jaehoon Lee and Kelvin Xu and Aaron T Parisi and Abhishek Kumar and Alexander A Alemi and Alex Rizkowsky and Azade Nova and Ben Adlam and Bernd Bohnet and Gamaleldin Fathy Elsayed and Hanie Sedghi and Igor Mordatch and Isabelle Simpson and Izzeddin Gur and Jasper Snoek and Jeffrey Pennington and Jiri Hron and Kathleen Kenealy and Kevin Swersky and Kshiteej Mahajan and Laura A Culp and Lechao Xiao and Maxwell Bileschi and Noah Constant and Roman Novak and Rosanne Liu and Tris Warkentin and Yamini Bansal and Ethan Dyer and Behnam Neyshabur and Jascha Sohl-Dickstein and Noah Fiedel},
journal={Transactions on Machine Learning Research},
issn={2835-8856},
year={2024},
url={https://openreview.net/forum?id=lNAyUngGFK},
note={Expert Certification}
}

@article{nivel2013bounded,
  title={Bounded recursive self-improvement},
  author={Nivel, Eric and Th{\'o}risson, Kristinn R and Steunebrink, Bas R and Dindo, Haris and Pezzulo, Giovanni and Rodriguez, Manuel and Hern{\'a}ndez, Carlos and Ognibene, Dimitri and Schmidhuber, J{\"u}rgen and Sanz, Ricardo and others},
  journal={arXiv preprint arXiv:1312.6764},
  year={2013}
}

@article{samuel1959some,
  title={Some studies in machine learning using the game of checkers},
  author={Samuel, Arthur L},
  journal={IBM Journal of research and development},
  volume={3},
  number={3},
  pages={210--229},
  year={1959},
  publisher={IBM}
}

@book{sutton1998reinforcement,
  title={Reinforcement learning: An introduction},
  author={Sutton, Richard S and Barto, Andrew G and others},
  volume={1},
  year={1998},
  publisher={MIT press Cambridge}
}

@inproceedings{arjovsky2017wasserstein,
  title={Wasserstein generative adversarial networks},
  author={Arjovsky, Martin and Chintala, Soumith and Bottou, L{\'e}on},
  booktitle={International conference on machine learning},
  pages={214--223},
  year={2017},
  organization={Pmlr}
}

@inproceedings{
zhou2026selfchallenging,
title={Self-Challenging Language Model Agents},
author={Yifei Zhou and Sergey Levine and Jason E Weston and Xian Li and Sainbayar Sukhbaatar},
booktitle={The Thirty-ninth Annual Conference on Neural Information Processing Systems},
year={2026},
url={https://openreview.net/forum?id=9yusqX9DpR}
}

@article{kulikov2026autodata,
  title   = "Autodata: an automatic data scientist to create high quality data",
  author  = {Kulikov, Ilia and Whitehouse, Chenxi and Wu, Tianhao and Saha, Swarnadeep and  Helenowski, Eryk and Yuan, Weizhe and Golovneva, Olga and Lanchantin, Jack and Bachrach, Yoram and Foerster, Jakob and Li, Xian and Fang, Han and Sukhbaatar, Sainbayar and Weston, Jason},
  year    = "2026",
  month   = "April",
  url     = "https://facebookresearch.github.io/RAM/blogs/autodata/"
}

@inproceedings{
sukhbaatar2018intrinsic,
title={Intrinsic Motivation and Automatic Curricula via Asymmetric Self-Play},
author={Sainbayar Sukhbaatar and Zeming Lin and Ilya Kostrikov and Gabriel Synnaeve and Arthur Szlam and Rob Fergus},
booktitle={International Conference on Learning Representations},
year={2018},
url={https://openreview.net/forum?id=SkT5Yg-RZ},
}

@inproceedings{agentgym,
    title = "{A}gent{G}ym: Evaluating and Training Large Language Model-based Agents across Diverse Environments",
    author = "Xi, Zhiheng  and
      Ding, Yiwen  and
      Chen, Wenxiang  and
      Hong, Boyang  and
      Guo, Honglin  and
      Wang, Junzhe  and
      Guo, Xin  and
      Yang, Dingwen  and
      Liao, Chenyang  and
      He, Wei  and
      Gao, Songyang  and
      Chen, Lu  and
      Zheng, Rui  and
      Zou, Yicheng  and
      Gui, Tao  and
      Zhang, Qi  and
      Qiu, Xipeng  and
      Huang, Xuanjing  and
      Wu, Zuxuan  and
      Jiang, Yu-Gang",
    editor = "Che, Wanxiang  and
      Nabende, Joyce  and
      Shutova, Ekaterina  and
      Pilehvar, Mohammad Taher",
    booktitle = "Proceedings of the 63rd Annual Meeting of the Association for Computational Linguistics (Volume 1: Long Papers)",
    month = jul,
    year = "2025",
    address = "Vienna, Austria",
    publisher = "Association for Computational Linguistics",
    url = "https://aclanthology.org/2025.acl-long.1355/",
    doi = "10.18653/v1/2025.acl-long.1355",
    pages = "27914--27961",
    ISBN = "979-8-89176-251-0",
}

@InProceedings{discover-loss-lu-2,
  title = 	 {Model-Free Opponent Shaping},
  author =       {Lu, Christopher and Willi, Timon and De Witt, Christian A Schroeder and Foerster, Jakob},
  booktitle = 	 {Proceedings of the 39th International Conference on Machine Learning},
  pages = 	 {14398--14411},
  year = 	 {2022},
  editor = 	 {Chaudhuri, Kamalika and Jegelka, Stefanie and Song, Le and Szepesvari, Csaba and Niu, Gang and Sabato, Sivan},
  volume = 	 {162},
  series = 	 {Proceedings of Machine Learning Research},
  month = 	 {17--23 Jul},
  publisher =    {PMLR},
  url = 	 {https://proceedings.mlr.press/v162/lu22d.html},
}

@InProceedings{ail-ghasemipour20a,
  title = 	 {A Divergence Minimization Perspective on Imitation Learning Methods},
  author =       {Ghasemipour, Seyed Kamyar Seyed and Zemel, Richard and Gu, Shixiang},
  booktitle = 	 {Proceedings of the Conference on Robot Learning},
  pages = 	 {1259--1277},
  year = 	 {2020},
  editor = 	 {Kaelbling, Leslie Pack and Kragic, Danica and Sugiura, Komei},
  volume = 	 {100},
  series = 	 {Proceedings of Machine Learning Research},
  month = 	 {30 Oct--01 Nov},
  publisher =    {PMLR},
  url = 	 {https://proceedings.mlr.press/v100/ghasemipour20a.html},
}

@misc{autoresearch,
  author       = {Karpathy, Andrej},
  title        = {autoresearch},
  year         = {2026},
  month        = mar,
  howpublished = {\url{https://github.com/karpathy/autoresearch}},
  note         = {GitHub repository}
}

@inproceedings{DBLP:conf/gecco/NasirETJC24,
  author       = {Muhammad Umair Nasir and
                  Sam Earle and
                  Julian Togelius and
                  Steven James and
                  Christopher W. Cleghorn},
  editor       = {Xiaodong Li and
                  Julia Handl},
  title        = {LLMatic: Neural Architecture Search Via Large Language Models And
                  Quality Diversity Optimization},
  booktitle    = {Proceedings of the Genetic and Evolutionary Computation Conference,
                  {GECCO} 2024, Melbourne, VIC, Australia, July 14-18, 2024},
  publisher    = {{ACM}},
  year         = {2024},
  url          = {https://doi.org/10.1145/3638529.3654017},
  doi          = {10.1145/3638529.3654017},
  bibsource    = {dblp computer science bibliography, https://dblp.org}
}

@inproceedings{guo2024connecting,
title={Connecting Large Language Models with Evolutionary Algorithms Yields Powerful Prompt Optimizers},
author={Qingyan Guo and Rui Wang and Junliang Guo and Bei Li and Kaitao Song and Xu Tan and Guoqing Liu and Jiang Bian and Yujiu Yang},
booktitle={The Twelfth International Conference on Learning Representations},
year={2024},
url={https://openreview.net/forum?id=ZG3RaNIsO8}
}

@inproceedings{agrawal2026gepa,
title={{GEPA}: Reflective Prompt Evolution Can Outperform Reinforcement Learning},
author={Lakshya A Agrawal and Shangyin Tan and Dilara Soylu and Noah Ziems and Rishi Khare and Krista Opsahl-Ong and Arnav Singhvi and Herumb Shandilya and Michael J Ryan and Meng Jiang and Christopher Potts and Koushik Sen and Alex Dimakis and Ion Stoica and Dan Klein and Matei Zaharia and Omar Khattab},
booktitle={The Fourteenth International Conference on Learning Representations},
year={2026},
url={https://openreview.net/forum?id=RQm2KQTM5r}
}

@inproceedings{DBLP:conf/iclr/HazraSPLM25,
  author       = {Rishi Hazra and
                  Alkis Sygkounas and
                  Andreas Persson and
                  Amy Loutfi and
                  Pedro Zuidberg Dos Martires},
  title        = {REvolve: Reward Evolution with Large Language Models using Human Feedback},
  booktitle    = {The Thirteenth International Conference on Learning Representations,
                  {ICLR} 2025, Singapore, April 24-28, 2025},
  publisher    = {OpenReview.net},
  year         = {2025},
  url          = {https://openreview.net/forum?id=cJPUpL8mOw},
  bibsource    = {dblp computer science bibliography, https://dblp.org}
}

@inproceedings{DBLP:conf/ijcnlp/XieHZYCLLGK25,
  author       = {Chulin Xie and
                  Yangsibo Huang and
                  Chiyuan Zhang and
                  Da Yu and
                  Xinyun Chen and
                  Bill Yuchen Lin and
                  Bo Li and
                  Badih Ghazi and
                  Ravi Kumar},
  editor       = {Kentaro Inui and
                  Sakriani Sakti and
                  Haofen Wang and
                  Derek F. Wong and
                  Pushpak Bhattacharyya and
                  Biplab Banerjee and
                  Asif Ekbal and
                  Tanmoy Chakraborty and
                  Dhirendra Pratap Singh},
  title        = {On Memorization of Large Language Models in Logical Reasoning},
  booktitle    = {Proceedings of the 14th International Joint Conference on Natural
                  Language Processing and the 4th Conference of the Asia-Pacific Chapter
                  of the Association for Computational Linguistics, {IJCNLP-AACL} 2025,
                  Mumbai, India, December 20-24, 2025},
  pages        = {2742--2785},
  publisher    = {The Asian Federation of Natural Language Processing and The Association
                  for Computational Linguistics},
  year         = {2025},
  url          = {https://aclanthology.org/2025.ijcnlp-long.148/},
  bibsource    = {dblp computer science bibliography, https://dblp.org}
}

@inproceedings{
valmeekam2022large,
title={Large Language Models Still Can't Plan (A Benchmark for {LLM}s on Planning and Reasoning about Change)},
author={Karthik Valmeekam and Alberto Olmo and Sarath Sreedharan and Subbarao Kambhampati},
booktitle={NeurIPS 2022 Foundation Models for Decision Making Workshop},
year={2022},
url={https://openreview.net/forum?id=wUU-7XTL5XO}
}

@inproceedings{DBLP:conf/nips/DziriLSLJLWWB0H23,
  author       = {Nouha Dziri and
                  Ximing Lu and
                  Melanie Sclar and
                  Xiang Lorraine Li and
                  Liwei Jiang and
                  Bill Yuchen Lin and
                  Sean Welleck and
                  Peter West and
                  Chandra Bhagavatula and
                  Ronan Le Bras and
                  Jena D. Hwang and
                  Soumya Sanyal and
                  Xiang Ren and
                  Allyson Ettinger and
                  Za{\"{\i}}d Harchaoui and
                  Yejin Choi},
  editor       = {Alice Oh and
                  Tristan Naumann and
                  Amir Globerson and
                  Kate Saenko and
                  Moritz Hardt and
                  Sergey Levine},
  title        = {Faith and Fate: Limits of Transformers on Compositionality},
  booktitle    = {Advances in Neural Information Processing Systems 36: Annual Conference
                  on Neural Information Processing Systems 2023, NeurIPS 2023, New Orleans,
                  LA, USA, December 10 - 16, 2023},
  year         = {2023},
  url          = {http://papers.nips.cc/paper\_files/paper/2023/hash/deb3c28192f979302c157cb653c15e90-Abstract-Conference.html},
  bibsource    = {dblp computer science bibliography, https://dblp.org}
}

@article{DBLP:journals/corr/abs-2506-06941,
  author       = {Parshin Shojaee and
                  Iman Mirzadeh and
                  Keivan Alizadeh and
                  Maxwell Horton and
                  Samy Bengio and
                  Mehrdad Farajtabar},
  title        = {The Illusion of Thinking: Understanding the Strengths and Limitations
                  of Reasoning Models via the Lens of Problem Complexity},
  journal      = {CoRR},
  volume       = {abs/2506.06941},
  year         = {2025},
  url          = {https://doi.org/10.48550/arXiv.2506.06941},
  doi          = {10.48550/ARXIV.2506.06941},
  eprinttype   = {arXiv},
  eprint       = {2506.06941},
  bibsource    = {dblp computer science bibliography, https://dblp.org}
}

@inproceedings{DBLP:conf/iclr/MirzadehASTBF25,
  author       = {Iman Mirzadeh and
                  Keivan Alizadeh and
                  Hooman Shahrokhi and
                  Oncel Tuzel and
                  Samy Bengio and
                  Mehrdad Farajtabar},
  title        = {GSM-Symbolic: Understanding the Limitations of Mathematical Reasoning
                  in Large Language Models},
  booktitle    = {The Thirteenth International Conference on Learning Representations,
                  {ICLR} 2025, Singapore, April 24-28, 2025},
  publisher    = {OpenReview.net},
  year         = {2025},
  url          = {https://openreview.net/forum?id=AjXkRZIvjB},
  bibsource    = {dblp computer science bibliography, https://dblp.org}
}

@inproceedings{DBLP:conf/icml/GuinetODC24,
  author       = {Gauthier Guinet and
                  Behrooz Omidvar{-}Tehrani and
                  Anoop Deoras and
                  Laurent Callot},
  editor       = {Ruslan Salakhutdinov and
                  Zico Kolter and
                  Katherine A. Heller and
                  Adrian Weller and
                  Nuria Oliver and
                  Jonathan Scarlett and
                  Felix Berkenkamp},
  title        = {Automated Evaluation of Retrieval-Augmented Language Models with Task-Specific
                  Exam Generation},
  booktitle    = {Forty-first International Conference on Machine Learning, {ICML} 2024,
                  Vienna, Austria, July 21-27, 2024},
  series       = {Proceedings of Machine Learning Research},
  pages        = {16773--16801},
  publisher    = {{PMLR} / OpenReview.net},
  year         = {2024},
  url          = {https://proceedings.mlr.press/v235/guinet24a.html},
  bibsource    = {dblp computer science bibliography, https://dblp.org}
}

@inproceedings{DBLP:conf/nips/MullerKB024,
  author       = {Luis M{\"{u}}ller and
                  Daniel Kusuma and
                  Blai Bonet and
                  Christopher Morris},
  editor       = {Amir Globersons and
                  Lester Mackey and
                  Danielle Belgrave and
                  Angela Fan and
                  Ulrich Paquet and
                  Jakub M. Tomczak and
                  Cheng Zhang},
  title        = {Towards Principled Graph Transformers},
  booktitle    = {Advances in Neural Information Processing Systems 38: Annual Conference
                  on Neural Information Processing Systems 2024, NeurIPS 2024, Vancouver,
                  BC, Canada, December 10 - 15, 2024},
  year         = {2024},
  url          = {http://papers.nips.cc/paper\_files/paper/2024/hash/e5419147e53eba322cf12aff266a66f2-Abstract-Conference.html},
  bibsource    = {dblp computer science bibliography, https://dblp.org}
}

@inproceedings{nie-etal-2020-adversarial,
    title = "Adversarial {NLI}: A New Benchmark for Natural Language Understanding",
    author = "Nie, Yixin  and
      Williams, Adina  and
      Dinan, Emily  and
      Bansal, Mohit  and
      Weston, Jason  and
      Kiela, Douwe",
    editor = "Jurafsky, Dan  and
      Chai, Joyce  and
      Schluter, Natalie  and
      Tetreault, Joel",
    booktitle = "Proceedings of the 58th Annual Meeting of the Association for Computational Linguistics",
    month = jul,
    year = "2020",
    address = "Online",
    publisher = "Association for Computational Linguistics",
    url = "https://aclanthology.org/2020.acl-main.441/",
    doi = "10.18653/v1/2020.acl-main.441",
    pages = "4885--4901"
}

@article{DBLP:journals/corr/abs-2506-13131,
  author       = {Alexander Novikov and
                  Ng{\^{a}}n Vu and
                  Marvin Eisenberger and
                  Emilien Dupont and
                  Po{-}Sen Huang and
                  Adam Zsolt Wagner and
                  Sergey Shirobokov and
                  Borislav Kozlovskii and
                  Francisco J. R. Ruiz and
                  Abbas Mehrabian and
                  M. Pawan Kumar and
                  Abigail See and
                  Swarat Chaudhuri and
                  George Holland and
                  Alex Davies and
                  Sebastian Nowozin and
                  Pushmeet Kohli and
                  Matej Balog},
  title        = {AlphaEvolve: {A} coding agent for scientific and algorithmic discovery},
  journal      = {CoRR},
  volume       = {abs/2506.13131},
  year         = {2025},
  url          = {https://doi.org/10.48550/arXiv.2506.13131},
  doi          = {10.48550/ARXIV.2506.13131},
  eprinttype   = {arXiv},
  eprint       = {2506.13131},
  bibsource    = {dblp computer science bibliography, https://dblp.org}
}

@article{funsearch,
  author       = {Bernardino Romera{-}Paredes and
                  Mohammadamin Barekatain and
                  Alexander Novikov and
                  Matej Balog and
                  M. Pawan Kumar and
                  Emilien Dupont and
                  Francisco J. R. Ruiz and
                  Jordan S. Ellenberg and
                  Pengming Wang and
                  Omar Fawzi and
                  Pushmeet Kohli and
                  Alhussein Fawzi},
  title        = {Mathematical discoveries from program search with large language models},
  journal      = {Nat.},
  volume       = {625},
  number       = {7995},
  pages        = {468--475},
  year         = {2024},
  url          = {https://doi.org/10.1038/s41586-023-06924-6},
  doi          = {10.1038/S41586-023-06924-6},
  bibsource    = {dblp computer science bibliography, https://dblp.org}
}

@inproceedings{nora,
  author       = {Anirban Das and
                  Irtaza Khalid and
                  Rafael Pe{\~{n}}aloza and
                  Steven Schockaert},
  title        = {When No Paths Lead to Rome: Benchmarking Systematic Neural Relational
                  Reasoning},
 booktitle    = {NeurIPS},
 year         = {2025}
}

@inproceedings{DBLP:conf/ijcai/MeilickeCRS19,
  author       = {Christian Meilicke and
                  Melisachew Wudage Chekol and
                  Daniel Ruffinelli and
                  Heiner Stuckenschmidt},
  editor       = {Sarit Kraus},
  title        = {Anytime Bottom-Up Rule Learning for Knowledge Graph Completion},
  booktitle    = {Proceedings of the Twenty-Eighth International Joint Conference on
                  Artificial Intelligence, {IJCAI} 2019, Macao, China, August 10-16,
                  2019},
  pages        = {3137--3143},
  publisher    = {ijcai.org},
  year         = {2019},
  url          = {https://doi.org/10.24963/ijcai.2019/435},
  doi          = {10.24963/IJCAI.2019/435},
  bibsource    = {dblp computer science bibliography, https://dblp.org}
}

@inproceedings{DBLP:conf/nips/BordesUGWY13,
  author       = {Antoine Bordes and
                  Nicolas Usunier and
                  Alberto Garc{\'{\i}}a{-}Dur{\'{a}}n and
                  Jason Weston and
                  Oksana Yakhnenko},
  editor       = {Christopher J. C. Burges and
                  L{\'{e}}on Bottou and
                  Zoubin Ghahramani and
                  Kilian Q. Weinberger},
  title        = {Translating Embeddings for Modeling Multi-relational Data},
  booktitle    = {Advances in Neural Information Processing Systems 26: 27th Annual
                  Conference on Neural Information Processing Systems 2013. Proceedings
                  of a meeting held December 5-8, 2013, Lake Tahoe, Nevada, United States},
  pages        = {2787--2795},
  year         = {2013},
  url          = {https://proceedings.neurips.cc/paper/2013/hash/1cecc7a77928ca8133fa24680a88d2f9-Abstract.html},
  bibsource    = {dblp computer science bibliography, https://dblp.org}
}

@article{kingma2017adam,
      title={Adam: A Method for Stochastic Optimization}, 
      author={Diederik P. Kingma and Jimmy Ba},
      year={2017},
      eprint={1412.6980},
}

@inproceedings{edge-transformer,
  author       = {Leon Bergen and
                  Timothy J. O'Donnell and
                  Dzmitry Bahdanau},
  editor       = {Marc'Aurelio Ranzato and
                  Alina Beygelzimer and
                  Yann N. Dauphin and
                  Percy Liang and
                  Jennifer Wortman Vaughan},
  title        = {Systematic Generalization with Edge Transformers},
  booktitle    = {Advances in Neural Information Processing Systems 34: Annual Conference
                  on Neural Information Processing Systems 2021, NeurIPS 2021, December
                  6-14, 2021, virtual},
  pages        = {1390--1402},
  year         = {2021},
  url          = {https://proceedings.neurips.cc/paper/2021/hash/0a4dc6dae338c9cb08947c07581f77a2-Abstract.html},
  bibsource    = {dblp computer science bibliography, https://dblp.org}
}

@inproceedings{DBLP:conf/iclr/ChengAS23,
  author       = {Kewei Cheng and
                  Nesreen K. Ahmed and
                  Yizhou Sun},
  title        = {Neural Compositional Rule Learning for Knowledge Graph Reasoning},
  booktitle    = {The Eleventh International Conference on Learning Representations,
                  {ICLR} 2023, Kigali, Rwanda, May 1-5, 2023},
  publisher    = {OpenReview.net},
  year         = {2023},
  url          = {https://openreview.net/pdf?id=F8VKQyDgRVj},
  bibsource    = {dblp computer science bibliography, https://dblp.org}
}

@inproceedings{r5,
  author       = {Shengyao Lu and
                  Bang Liu and
                  Keith G. Mills and
                  Shangling Jui and
                  Di Niu},
  title        = {{R5:} Rule Discovery with Reinforced and Recurrent Relational Reasoning},
  booktitle    = {The Tenth International Conference on Learning Representations, {ICLR}
                  2022, Virtual Event, April 25-29, 2022},
  publisher    = {OpenReview.net},
  year         = {2022},
  url          = {https://openreview.net/forum?id=2eXhNpHeW6E},
  bibsource    = {dblp computer science bibliography, https://dblp.org}
}

@article{DBLP:journals/ml/CropperDEM22,
  author       = {Andrew Cropper and
                  Sebastijan Dumancic and
                  Richard Evans and
                  Stephen H. Muggleton},
  title        = {Inductive logic programming at 30},
  journal      = {Mach. Learn.},
  volume       = {111},
  number       = {1},
  pages        = {147--172},
  year         = {2022},
  url          = {https://doi.org/10.1007/s10994-021-06089-1},
  doi          = {10.1007/S10994-021-06089-1},
  bibsource    = {dblp computer science bibliography, https://dblp.org}
}

@inproceedings{DBLP:conf/icml/Minervini0SGR20,
  author       = {Pasquale Minervini and
                  Sebastian Riedel and
                  Pontus Stenetorp and
                  Edward Grefenstette and
                  Tim Rockt{\"{a}}schel},
  title        = {Learning Reasoning Strategies in End-to-End Differentiable Proving},
  booktitle    = {Proceedings of the 37th International Conference on Machine Learning,
                  {ICML} 2020, 13-18 July 2020, Virtual Event},
  series       = {Proceedings of Machine Learning Research},
  volume       = {119},
  pages        = {6938--6949},
  publisher    = {{PMLR}},
  year         = {2020},
  url          = {http://proceedings.mlr.press/v119/minervini20a.html},
  bibsource    = {dblp computer science bibliography, https://dblp.org}
}

@inproceedings{Sinha2019CLUTRR,
	title={CLUTRR: A diagnostic benchmark for inductive reasoning from text},
	author={Sinha, Koustuv and Sodhani, Shagun and Dong, Jin and Pineau, Joelle and Hamilton, William L.},
	booktitle={Proceedings of EMNLP-IJCNLP},
	pages={4505--4514},
	year={2019},
	publisher={Association for Computational Linguistics},
	doi={10.18653/V1/D19-1458},
	url={https://doi.org/10.18653/v1/D19-1458}
}

@inproceedings{Cohen2019GraphLog,
	author    = {Cohen, William W.},
	title     = {GraphLog: A Benchmark for Logical Learning on Graphs},
	booktitle = {Proceedings of the 2019 International Symposium on Inductive Logic Programming},
	year      = {2019}
}

@article{gebser2011potassco,
	title={Potassco: The Potsdam answer set solving collection},
	author={Gebser, Martin and Kaufmann, Benjamin and Kaminski, Roland and Ostrowski, Max and Schaub, Torsten and Schneider, Marius},
	journal={Ai Communications},
	volume={24},
	number={2},
	pages={107--124},
	year={2011},
	publisher={SAGE Publications Sage UK: London, England}
}

@article{zhu2021neural,
	title={Neural {Bellman-Ford} networks: A general graph neural network framework for link prediction},
	author={Zhu, Zhaocheng and Zhang, Zuobai and Xhonneux, Louis-Pascal and Tang, Jian},
	journal={Advances in neural information processing systems},
	volume={34},
	pages={29476--29490},
	year={2021}
}

@article{khalid2025systematic,
	title={Systematic relational reasoning with epistemic graph neural networks},
	author={Khalid, Muhammad and Schockaert, Steven},
	year={2025},
		journal={The Thirteenth International Conference on Learning Representations
		},
}

@inproceedings{cheng2023neural,
  author       = {Kewei Cheng and
                  Nesreen K. Ahmed and
                  Yizhou Sun},
  title        = {Neural Compositional Rule Learning for Knowledge Graph Reasoning},
  booktitle    = {The Eleventh International Conference on Learning Representations,
                  {ICLR} 2023, Kigali, Rwanda, May 1-5, 2023},
  publisher    = {OpenReview.net},
  year         = {2023},
  url          = {https://openreview.net/forum?id=F8VKQyDgRVj},
  bibsource    = {dblp computer science bibliography, https://dblp.org}
}

@article{tarjan-1972,
author = {Tarjan, Robert},
title = {Depth-First Search and Linear Graph Algorithms},
journal = {SIAM Journal on Computing},
volume = {1},
number = {2},
pages = {146-160},
year = {1972},
doi = {10.1137/0201010},
URL = { https://doi.org/10.1137/0201010},
eprint = {  https://doi.org/10.1137/0201010},
}

%%%%%%%%%%%%%%%%%%%%%%%%%%%%%%%%%%%%%%%%%%%%%%%%%%%%%%%%%%%%

\appendix
\newpage
\section{Additional details}\label{appDetails}

\subsection{Background on Datalog}\label{appDetailsDatalog}
We rely on Datalog rules to encode the knowledge for reasoning about KGs. An example of a Datalog rule is as follows:
$$
\textit{father\_of}(X,Z) \leftarrow \textit{father\_of}(X,Y) \wedge \textit{brother\_of}(Y,Z)
$$
This rule encodes that whenever some entity $X$ is known to be the father of $Y$ and $Y$ is known to be the brother of $Z$, we can infer that $X$ is the father of $Z$. Following standard practice, we use uppercase arguments such as $X$ to denote variables and lowercase arguments such as \textit{bob} to denote specific entities (i.e.\ constants).  A literal is an expression of the form $r(\alpha_1,...,\alpha_n)$ where $r$ is a relation and $\alpha_1,...,\alpha_n$ are either constants or variables.
A Datalog rule is an expression of the form $\textit{head} \leftarrow \textit{body}$, where \textit{head} is a literal and \textit{body} is a conjunction of literals, such that every variable which appears in \textit{head} also appears in \textit{body}. In this paper, we will not be concerned with literals that have more than two arguments. If all arguments of the literal $\alpha$ are constants, we call $\alpha$ a ground literal. Similarly, a ground rule is a rule in which only ground literals occur. Given a set of Datalog rules $\mathcal{P}$ and a KG $\mathcal{G}$, the grounding $\mathcal{P}_{\mathcal{G}}$ is defined as the set of all ground rules that can be constructed from the rules in $\mathcal{P}$ by instantiating their variables by entities from $\mathcal{G}$.

\paragraph{Entailment}
Given a KG $\mathcal{G}$ and a set of Datalog rules $\mathcal{P}$, we write $\mathcal{G}\cup\mathcal{P} \models r(e,f)$ to denote that $r(e,f)$ can be inferred from $\mathcal{G}$ using the rules in $\mathcal{P}$. This entailment relation can be formally defined as follows. We write $\mathcal{G}\cup\mathcal{P} \models_0 r(e,f)$ to denote that $r(e,f)\in\mathcal{G}$. For $i>1$ we have $\mathcal{G}\cup\mathcal{P} \models_i r(e,f)$ iff $\mathcal{G}\cup\mathcal{P} \models_{i-1} r(e,f)$ or there is a rule $r(e,f) \leftarrow \alpha_1 \wedge ... \wedge \alpha_k$ in $\mathcal{P}_{\mathcal{G}}$ such that $\mathcal{G}\cup\mathcal{P} \models_{i-1} \alpha_j$ holds for every $j\in\{1,...,k\}$. We then have $\mathcal{G}\cup\mathcal{P} \models r(e,f)$ iff $\mathcal{G}\cup\mathcal{P} \models_i r(e,f)$ for some $i\in\mathbb{N}$.

\paragraph{Path-based and off-path reasoning}
When reasoning about knowledge graphs, so-called closed path rules often play an important role \citep{DBLP:conf/ijcai/MeilickeCRS19}. These are rules of the following form:
$$
r(X_1,X_n) \leftarrow r_1(X_1,X_2) \wedge ... \wedge r_{n-1}(X_{n-1},X_n)
$$
When all rules in $\mathcal{P}$ are closed path rules, then $\mathcal{G}\cup\mathcal{P} \models r(e,f)$ holds iff there is a path $\mathcal{G}_0=\{(e,r_1,e_1),(e_1,r_2,e_2),...,(e_{k-1},r_k,f)\}$ such that $\mathcal{G}_0\cup \mathcal{P} \models r(e,f)$. Methods such as NCRL \citep{DBLP:conf/iclr/ChengAS23} directly exploit this connection, which simplifies learning significantly. For instance, CLUTRR \citep{Sinha2019CLUTRR} only involves closed path rules, which is why NCRL performs well on this benchmark. Unfortunately, path-based methods are not sufficient for reasoning about other types of rules, as illustrated in the next example. 

\begin{example}\label{exOffPathreasoning}
The rule base $\mathcal{P}_{\textit{nora}}$ from NoRA \citep{nora} includes the following rules:
\begin{align*}
\textit{has\_property}(X, \textit{no\_daughters}) &\leftarrow \textit{parent\_of}(X,Y) \wedge \textit{belongs\_to\_group}(Y, \textit{male})\\
&\quad\quad\quad\wedge \textit{has\_property}(Y, \textit{no\_sisters})\\
\textit{belongs\_to\_group}(Y, \textit{male}) &\leftarrow \textit{parent\_of}(X,Y) \wedge \textit{has\_property}(X, \textit{no\_daughters})\\
\textit{father\_of}(X,Y) &\leftarrow \textit{parent\_of}(X,Y) \wedge \textit{belongs\_to\_group}(X, \textit{male})
\end{align*}
Let us furthermore consider the following KG:
\begin{align*}
\mathcal{G} &= \{\textit{parent\_of}(\textit{bob},\textit{dave}),\allowbreak \textit{belongs\_to\_group}(\textit{dave}, \textit{male}),\allowbreak \textit{has\_property}(\textit{dave}, \textit{no\_sisters}),\\
&\quad\quad \textit{parent\_of}(\textit{bob},\textit{john}),\allowbreak \textit{parent\_of}(\textit{john},\textit{alice})\}
\end{align*}
We have $\mathcal{G}\cup \mathcal{\mathcal{P}_{\textit{nora}}} \models \textit{father\_of}(\textit{john},\textit{alice})$, but this inference relies on knowledge about \textit{dave} (to infer that $\textit{bob}$ has no daughters), even though \textit{dave} is not on any path between \textit{john} and \textit{alice} in $\mathcal{G}$.
\end{example}

\subsection{Generating model-agnostic samplers using LLMs}\label{appGeneratingSamplersDetails}
The samplers from Section \ref{secModelAgnosticSamplers} were generated and iteratively refined by prompting Claude Opus 4.6 through a dialogue. The LLM was initially requested to create a new set of world rules\footnote{described in more details in Appendix \ref{appGeneratingWorldRulesDetails}} and then to design an associated sampler in the form of a Python program. Next, we prompted the model to generalize the approach, to make it suitable for any set of rules. Throughout this conversation, we tested the produced samplers by executing the code and collecting sets of story graphs for rule compliance checks. These story graphs were validated locally using the Clingo ASP solver\footnote{\url{https://potassco.org/clingo/}} \citep{gebser2011potassco}, to detect any violation of the rules in the sampled graphs. When errors were found, they were passed back to Claude, with a prompt requesting the model to identify the problem and apply the necessary corrections. It is worth noting that Claude Opus 4.6 does not have access to Clingo in its sandbox. However, it relies on forward derivation chains implemented in the samplers to explore the worlds, and run internal tests related to the correctness of the sampled stories, relying mostly on Python tool calls. The final version of each sampler was obtained after less than 10 update requests. More details about the generation process and the detected errors are given below.

\paragraph{First version of the samplers} The model was prompted to write Python programs that sample knowledge graphs from a set of rules (either LLM-generated or user provided). The total number of vertices in the sampled graphs is fixed in advance and passed to the Python program as an argument. We specified in the prompt that the objective was to sample graphs that are \textit{challenging for reasoning models} and would \textit{require systems to make difficult inferences beyond transitivity and hierarchy reasoning steps}. The model is asked to try and design a smart sampling strategy, which could be based on \textit{sampling one edge at a time or expanding a seed graph recursively with subgraph motifs}. We also indicated in the prompt that any creative method can be used. The first world-specific samplers were obtained in this way, including the first version of the Cl-N-temp sampler. The two other NoRa-specific samplers were obtained by prompting the model to design two further algorithms, different from this first sampler and different from each other. We thus obtained the first versions of the Cl-N-greedy and Cl-N-bw samplers. General samplers were obtained by prompting Claude to extend the specific samplers so that they can be used for any set of world rules. Claude Opus 4.6 generated the Cl-hill, Cl-bw, Cl-motif and Cl-atlas samplers in this way. 

\paragraph{Errors and refinements} Given the initial version of a sampler, we then asked the model to verify, edit and eventually improve the samplers. %and ensure that they sample various graphs several times. 
In general, we found the initial samplers to be quite accurate. Apart from Datalog rules, which is what we focus on in this paper, we also experimented with more general types of rules, in particular rules with negation-as-failure and choice rules. These more general types of rules posed more problems for the samplers.
% for the sets of rules without negation and choice rules, such as \textbf{NoRa} and \textbf{Iron Coast} that are used in this work. They struggle more to adapt to additional constraints that are left for future work. 
We observed several main types of errors when testing the samplers on the NoRa world:
\begin{itemize}
\item The arity of the relations was not immediately handled correctly. This was particularly the case for unary predicates expressed with two slots, such are $\textit{person}(a,a)$.

\item The required number of entities in the sampled KGs is specified as an argument to the sampler. The initial samplers did not always ensure the sampled graphs had the correct number of entities.

\item NoRA's 292 rules create circular dependencies between all 51 predicates. Cycle detection in the rules was not handled in the first versions of the samplers, leading several samplers, such as Cl-bw to be caught in infinite loops.

\end{itemize}

In addition to NoRA, the general samplers were tested on ten different sets of world rules, which were generated by different LLMs. 
%They were then refined based on feedback about the errors detected by a Clingo-based validation script, as for the NoRA-specific samplers.

\paragraph{Generated samplers}
We now describe the main strategy used by each of the generated samplers. 
\begin{itemize}
    \item \textbf{Cl-hill}: this hill-climbing algorithm first analyzes and categorizes the rules, using categories such as base, derived, seedable, symmetric, constraint predicates, join patterns from multi-body rules and transitivity rules. It then seeds an initial set of base facts. Finally, it relies on hill climbing to optimize graphs for deep derivations and fewer facts, by operating several mutations on the initial seed graph and scoring the mutated graphs. The possible mutations include the addition of an entire sub-graph at once, to improve graphs connectivity and add transitive chains. It implements a form of forward chaining to measure the associated proof depths.

    \item \textbf{Cl-bw}: this backward sampling method is goal-directed in the sense that it guarantees that every generated graph contains at least one deep derivation chain by construction. Its graph scoring function is similar to the one used by Cl-hill. It additionally implements Strongly Connected Component (SCC) analysis to detect and handle circular rules\footnote{The generated script suggests that it relies on the algorithm from \cite{tarjan-1972}.}. For each target predicate, the sampler traces backwards through the rules to build a \emph{proof skeleton}, i.e.\ a template proof tree with variables instead of constants. The proof templates are then instantiated with specific constants and the resulting skeletons are combined into graphs, with constants shared across skeletons so the proofs interconnect. A population of graphs is generated with this method, and the best graphs are selected according to a scoring function. Candidates are ranked first by a cheap proxy score, and the top few candidates are then evaluated with the full scoring function.

    \item \textbf{Cl-motif}: similarly to the backward sampler, and unlike the hill climbing sampler, this method has a certain level of rule awareness. Motifs are similar to the proof skeletons from Cl-bw, but at a smaller scale. They are small self-contained patterns that trigger specific derivation rules. They are instantiated with facts and used as building blocks with shared variables to obtain complete graphs. Generated graphs are then ranked with various strategies for a selection of the best ones. This sampler's implementation is general in the sense that is can be used with new sets of rules, but it also contains a few heuristics which are specific to the worlds that were used during the validation and refinement process.

    \item \textbf{Cl-atlas}: this hybrid sampler combines ideas Cl-hill, Cl-bw and Cl-motif. In a first step, a ``strategy mix'' is chosen at random among four weighted items. Each strategy mix combines two elements such as ``backward + joins'' or ``joins + chains''. Once graphs are built following a given strategy mix, hill climbing is applied to improve it by trying random mutations. Cl-Atlas is also the only sampler where we specifically asked, in the Claude prompt, to design a method that would defeat a GNN evaluator. In consequence, (i) the graph scoring function explicitly favors deep reasoning chains, because maximum derivation depth is said to correspond to the number of message-passing steps a graph neural network would need; (ii) predicate diversity is expected to force multi-relational reasoning rather than single-relation pattern matching; and (iii) distractors (irrelevant facts that look useful but are not in any proof) are added because they are expected to make the GNN's discrimination harder.
    
\item \textbf{Cl-N-temp}: with this NoRA-specific method, graphs are first generated using a few family skeleton templates (e.g. \textit{nuclear}, \textit{multi-generational} and \textit{all male children} families) and hard-coded fact selection heuristics such as strategic gender omission to trigger gender inference chains.
The best graphs are then selected using a score that estimates reasoning complexity from graph topology, based on the presence or number of certain types of predicates. For example, the score favours the presence of multiple underage children in the story because it triggers \textit{living\_in\_same\_place} queries, which typically occur in complex reasoning chains.

\item \textbf{Cl-N-greedy}: this NoRA-specific algorithms starts from a minimal two-person married seed. The graph is then expanded one edge at a time by scoring candidate mutations (e.g.\ \textit{add child}, \textit{add marriage}, \textit{flip gender}). Candidate mutations are scored using forward chaining, to measure the inference depth and the number of non-trivial inferences. The method greedily selects the mutation that maximizes this composite score.

\item \textbf{Cl-N-bw}: this is a NoRA-specific variant of the general backward sampler, with an identical long rule chain sampling goal, but where heuristics specific to the NoRA world are hard-coded. In particular, the algorithm relies on seven hard-coded proof recipes, targeting long derivations in NoRA, instead of building abstract proof skeletons from the available rules. Graphs are built by instantiating and combining proof trees, maximizing the number of deep derivations per entity. 
Gender information is hidden were possible, to make inference harder.
\end{itemize}

Note that Cl-N-greedy and Cl-hill are not rule-aware, in the sense that they randomly generate facts and then apply a scoring metric to decide if a given modification should be preserved. In contrast, the two backward samplers and the Cl-motif sampler explicitly rely on the rules and the proof derivations to construct the graphs. Several of the samplers also crucially rely on motifs. In addition to Cl-motif, Cl-hill and Cl-N-temp also took into account the motif hint that was given as part of the prompt. The motifs are implemented with different levels of generalization, from hard-coded templates to more abstract rule-aware skeletons. The Atlas sampler was the last generated sampler. It is the only sampler where the prompt explicitly referred to the task of defeating a GNN-based reasoner. As a result, the sampler relies on distractor edges, and tried to diversify the generated graphs by a random strategy selection initial step. However, all the samplers rely on a similar scoring function,  which makes assumptions about what makes a graph challenging for reasoning models, based on known difficulties such as long rule chains. Unlike the evolutionary algorithm, the scoring function is blind to the real bottlenecks encountered by the ET model.

\subsection{LLM-based improvement of priority functions}\label{appImprovingPriorityFunctionsDetails}

\paragraph{FunSearch Prompt} The LLM is requested to output a priority function that is an improvement over two previously generated priority functions. The prompt contains a full explanation of the evolutionary experimental setting, a detailed description of ASP syntax and the task. A set of callable graph tools are also provided under the form of a list of Python functions. Final format instructions and two previously generated priority functions are included in the body of the prompt.

\newpage
\begin{tcolorbox}[colback=teal!5!white,colframe=teal!75!black,title=\textsc{Summary of the prompt for the generation of new priority functions}, width=\linewidth, nobeforeafter]
\footnotesize

\textbf{\textsc{Instructions}}

\vspace{2mm}

You are an AI participating in an evolutionary algorithm search to generate candidate priority functions.

\vspace{2mm}

\textsc{Task overview}

\vspace{2mm}
You are designing a function:

\begin{lstlisting}
priority(cand_fact, definite_rules_program, entailed_facts, facts_program)
    -> float
\end{lstlisting}

The purpose of this function is to score how useful a candidate edge would be if added to an existing graph, given what additional edges become entailed in that graph by the definite\_rules\_program.

\vspace{4mm}

\textsc{Evolutionary progression}

\vspace{2mm}

 priority\_v0 and a complete priority\_v1 are shown. priority\_v1 is already an improvement over priority\_v0. priority\_v2 is incomplete. Your job is to output a complete priority\_v2 that improves on priority\_v1.

\vspace{4mm}

\textsc{Arguments}
\begin{itemize}
\item \lstinline|cand_fact (str)|: this is a single directed edge string you are scoring. It is exactly one edge statement ending in a period.
\item \lstinline|definite_rules_program (str)|: this is a string containing many graph-transformation rules, one rule per line. These rules describe when the presence of certain edge patterns in the graph implies that another edge must also exist.
\item \lstinline|entailed_facts (str)|: this is a string containing directed edges that become implied after adding cand\_fact to facts\_program and then applying the graph-transformation rules, excluding edges that were already explicitly present in facts\_program (and excluding the candidate edge itself).
\item  \lstinline|facts_program (str)|: this is the existing graph, a string containing directed edges already present, usually one per line, each ending in a period.

\end{itemize}

\vspace{4mm}

\textbf{\textsc{Graph toolkit}}

\vspace{2mm}

You may call pre-built graph-theoretic functions directly inside your priority function — no import statement is needed, they are automatically available. The available functions are:

\vspace{2mm}
\begin{lstlisting}
# --- Path Analysis ---
shortest_path_length(G, src, tgt) -> int | None
    Number of hops on the shortest directed path. Returns None if unreachable.
...
# --- Structure ---
node_degree(G, node, mode='all') -> int
    Degree of a node. mode: 'in' (incoming), 'out' (outgoing), 'all' (both).
...
\end{lstlisting}

\textbf{\textsc{Evolution strategies}}

\vspace{2mm}

If you only see complete priority\_v0, try simple approaches first:
\begin{itemize}
\item Count implied edges
\item Find candidate edges whose endpoint nodes are common with nodes already appearing in facts\_program
\item Return any positive number based on simple metrics
\end{itemize}

When you see full demonstrations priority\_v0, priority\_v1, and need to complete priority\_v2:
\begin{enumerate}
    \item \textsc{Tweak the best}: Take priority\_v{i} (the current best) and make small modifications:
\begin{itemize}
   \item Adjust constants/weights slightly (e.g., change 0.5 to 0.7, or 2.0 to 1.5)
   \item Add a small bonus/penalty term to the existing logic
   \item Modify thresholds or conditions slightly
   \item ...
\end{itemize}

\item \textsc{Try something completely new}: Ignore previous versions and experiment with a fresh idea
\end{enumerate}

\textbf{\textsc{The two priority functions}}

\vspace{2mm}
\{priority\_v0\}

\{priority\_v1\}

\end{tcolorbox}

\subsection{Auto-Research Agent-in-the-loop initiation prompt}\label{appAutoresearchPrompt}
\begin{tcolorbox}[colback=teal!5!white,colframe=teal!75!black,title=\textsc{Program.md}, width=\linewidth, nobeforeafter]
\footnotesize
\footnotesize

\textbf{Experimental loop (agent-in-the-loop).} 

An outer loop over rounds where a coding agent iterates on a single \texttt{priority(...)} function per round, then retrains and evaluates an Edge Transformer (ET) end-to-end.

\textbf{Goal and criteria.} 

Write a priority function that drives the ET's eval accuracy \emph{down}; lower is better. All else being equal, simpler is better: a small improvement that adds ugly complexity is not worth it; removing code while matching or improving accuracy is a simplification win. Weigh complexity cost against improvement magnitude --- a $0.001$ gain from twenty new lines of hacky code is rejected; a $0.001$ gain from \emph{deleting} code is kept; equal accuracy with simpler code is kept. The first run always establishes the baseline.

\textbf{Key idea.} 

One round $=$ one priority function (\texttt{priority\_fns/round\_N.py}). The same function is used to generate \texttt{eval.csv} (held-out via seed offset), generate \texttt{base\_train.csv}, train a new ET on the merged \texttt{final\_train.csv}, and evaluate the new ET (also backfilling the previous ET on the new round's \texttt{eval.csv}). \emph{No internal proposal loop} runs over multiple priority functions inside a round; FunSearch / ProgramDB evolution is intentionally removed.

\textbf{Per-iteration agent loop.} 

For round $N$: (1) \emph{read} previous-round artifacts (metrics, sample data); (2) \emph{decide} a change to \texttt{priority(...)} (heuristic tweak, new signal, etc.); (3) \emph{write} \texttt{priority\_fns/round\_N.py}; (4) \emph{run} exactly one round of the pipeline.

\textbf{Round artifacts.} 

All outputs live under \texttt{RUN\_DIR/round\_N/}: \texttt{best\_priority\_fn.py} (exact code used), \texttt{eval.csv}, \texttt{base\_train.csv}, \texttt{final\_train.csv}, \texttt{model.pth}, and \texttt{eval\_summary.csv} (covering this round's eval, the previous round's eval, the base train split, and the merged train split). Run-level \texttt{records.pkl} stores per-ET metric dicts including \texttt{acc\_roundN\_eval}, \texttt{acc\_roundN-1\_eval}, \texttt{acc\_roundN\_base\_train}, \texttt{acc\_roundN\_final\_train}, and \texttt{acc\_roundN+1\_eval} (backfill: previous ET on the new \texttt{eval.csv}).

\textbf{Workflow.} 

(i) Activate the environment; (ii) create round $N$'s priority file by copying round $N-1$ and editing it; (iii) run one round end-to-end via \texttt{scripts/run\_multiround.py} with \texttt{-{}-init\_model\_path} pointing at the pretrained ET\textsubscript{0}; (iv) read the round's \texttt{eval\_summary.csv}, \texttt{records.pkl}, \texttt{eval.csv}, and \texttt{best\_priority\_fn.py}; (v) propose the next priority function. Reusing \texttt{-{}-run\_dir} naively recreates \texttt{round\_1}; strict incremental rounds in one run dir would need an explicit resume mode (not implemented).

\textbf{Parameters and defaults.} 

\texttt{run\_multiround.py} reads defaults from a single FullFlow JSON config: \texttt{multi\_round.*} for training and eval knobs (story counts, ET hyperparameters, merge cap), and \texttt{evaluation.*} for base story-generation knobs (entity range, base seed).

\textbf{Main idea.} 

The agent is a completely autonomous researcher: try things, keep what works, move on from what doesn't, rewind only very sparingly. \emph{Never stop.} Once the loop has begun, the agent does not pause to ask whether to continue --- it does not ask ``should I keep going?'' or ``is this a good stopping point?''. The human may be away, and the agent is expected to keep working until the configured number of rounds is complete or it is manually interrupted. If it runs out of ideas, it thinks harder: re-reads referenced papers, re-reads in-scope files for new angles, combines previous near-misses, attempts more radical changes. The loop runs until the human interrupts it.
\end{tcolorbox}

\subsection{Generating world rules using LLMs}\label{appGeneratingWorldRulesDetails}
LLMs were prompted to generate world rules that can be used to then create challenging reasoning problems. Our prompt describes the ASP syntax, presents the expected rule structure and provide examples. It also contains a few hints regarding the expected characteristics of challenging sets of rules (e.g.\ the fact that the rule bases should pose reasoning challenges beyond hierarchies and transitivity). The prompt presented below was iteratively refined by querying several models, namely GPT-5.3, Claude Sonnet 4.6 and Gemini 3.1 Pro. Constraints, such as the minimum number of rules to include in the final output and further format restrictions, were progressively added to the prompt during this exploration step, to ensure that the created worlds are complex enough and the syntax of their rules matches the our samplers' parsers.

We additionally noticed that passing the NoRA rules to the LLMs, after giving these instructions, and asking them to create a new world that is both different from NoRA and more challenging, seemed to lead to more complex and interesting sets of rules. The \textbf{Iron Coast} world was obtained with this approach, using Claude Opus 4.6 with our refined prompt. Additional instructions that were used for the creation of this world are to exclude negations and choice rules. Although our approach can in principle be extended to a richer logical syntax, this is beyond the scope of the current paper.

\newpage
\begin{tcolorbox}[colback=teal!5!white,colframe=teal!75!black,title=\textsc{Prompt for the generation of world rules}, width=\linewidth, nobeforeafter]
\footnotesize
\vspace{2mm}
Your role is to generate new sets of rules used to construct graphs that pose challenging reasoning problems. Your final response must contain a list of domain base facts and the entire set of created rules.

\vspace{2mm}
I'm looking for an interesting rule base that can be used for evaluating the reasoning abilities of AI models. The rule base should be encoded using Answer Set Programming (ASP) syntax. In particular, it can contain the following types of rules.
\vspace{2mm}

Facts have the following form.
$$
\text{r(a,b).}
$$

This fact states that the relationship r holds between the objects a and b. r(a,a) can also be written r(a).

\vspace{2mm}

Next, standard rules have the following form:
$$
\text{r(X,Z) :- u(X,Y), v(Y,Z).}
$$
r(X,X) can also be written r(X).

\vspace{2mm}
Here X,Y,Z are variables. The rule states that when X is in relation u with Y, and Y is in relation v with Z, then X is in relation r with Z. The number of conditions in a rule can vary. For instance, the following are also valid examples of rules:
$$
\text{r(X,Y) :- s(Y,X).}
$$
$$
\text{r(X,Y) :- u(X,Y), v(Y,Z).}
$$
In addition to standard rules, we also have choice rules, which have the following form:
$$
\text{\{r(X,Y),s(X,Y)\} :- u(X,Y).}
$$
This choice rule means that whenever X is in relation u with Y, then at least one of r(X,Y) or s(X,Y) should hold. Other examples of choice rules are:
$$
\text{r(X,Y) :- s(Y,X).}
$$
$$
\text{r(X,Y) :- u(X,Y), v(Y,Z).}
$$
Choice rules are typically used in combination with constraints. These take the following form:
$$
\text{:- r(X,Y), s(X,Y).}
$$
This constraint means that r(X,Y) and s(X,Y) cannot be jointly satisfied. Another example is:
$$
\text{:- r(X,Y), s(X,Z), Y!=Z.}
$$
This means that r(X,Y), and s(X,Z) cannot both be satisfied when Y is different from Z.

\vspace{4mm}

Additional constraints for this project :

\begin{itemize}
    \item Relations must not have more than two arguments.
    \item A relation must appear in your final answer only if it is used in more than one rule. Note that we are interested in the creation of rule bases that pose reasoning challenges beyond hierarchies and transitivity. The rules must be connected, and connected in various ways so that the reasoning challenges are diverse.
    \item Do not use any numerical value in the rules. The following example is not acceptable : 
    $$
    \text{:- reviewer(R), \#count { P : assigned(R, P) } > 2.}
    $$ 
    \item Create at least 40 different rules. 
    \item The rules should be preceded by a list of domain base facts.
    
\end{itemize}

\vspace{4mm}

These are good examples of rules, that are part of a rule base in the domain of family relations: 
$$
\text{maternal\_grandmother\_of(V,X) :- paternal\_grandmother\_of(U,X), grandmother\_of(V,X), U != V.} 
$$
$$
\text{is\_female(X) :- mother\_of(X,Y).}
$$
$$
\text{\{father(X,Y),mother(X,Y)\} :- parent(X,Y).}
$$

\vspace{4mm}

Please create an extensive rule base which describes some domain of interest and which is likely to be challenging for AI models. Check carefully the consistency of the rules you propose. Your final response must be a complete list of all the rules of the world you created and may contain comments on lines starting with "\%". 

\vspace{3mm}
\end{tcolorbox}

\begin{table}
\footnotesize
\centering
\caption{Cross-evaluation of samplers on Iron Coast in terms of exact-match accuracy. }\label{tabCrossEvaluationSamplersAtlas}
\begin{tabular}{llcccccc}
        \toprule
        && \multicolumn{6}{c}{\textbf{Test set}}\\
\cmidrule(lr){3-8}
         && \textbf{Cl-Atlas} & \textbf{Cl-bw} & \textbf{Cl-hill} & \textbf{Evo1} & \textbf{Evo2} & \textbf{Evo3}  \\
        \midrule
        \multirow{6}{*}{\rotatebox[origin=c]{90}{\textbf{Training set}} } & \textbf{Cl-Atlas} & - & \heat{0.01} & \heat{0.15} & \heat{0.03} & \heat{0.00} & \heat{0.02} \\
        &\textbf{Cl-bw} & \heat{0.03} & - & \heat{0.03}  & \heat{0.34} & \heat{0.20} & \heat{0.18} \\
        &\textbf{Cl-hill} & \heat{0.13} & \heat{0.02} & - & \heat{0.04} & \heat{0.00} & \heat{0.04}  \\
        &\textbf{Evo1} & \heat{0.10} & \heat{0.26} & \heat{0.07} & - & \heat{0.48} & \heat{0.28}  \\
        &\textbf{Evo2} & \heat{0.03} & \heat{0.23} & \heat{0.05} & \heat{0.34} & - & \heat{0.25}  \\
        &\textbf{Evo3} & \heat{0.06} & \heat{0.27} & \heat{0.06} & \heat{0.42} & \heat{0.47} & -  \\
        \bottomrule
    
    \end{tabular}
\end{table}

\begin{table}
\footnotesize
\centering
\caption{Cross-evaluation of priority functions obtained through evolutionary search (for NoRA), using three different models: qwen3-coder-next-80b (qw), gpt-oss-120b (gpt) and deepseek-coder-33b (ds). Results are reported in terms of exact-match accuracy. Each model was used in four independent runs. }\label{tabCrossEvaluationSamplersLLMcomparison}
\begin{tabular}{llcccccccccccc}
\toprule
&& \multicolumn{12}{c}{\textbf{Test set}}\\
\cmidrule(lr){3-14}
& & \textbf{qw1} & \textbf{qw2} & \textbf{qw3} & \textbf{qw4} & \textbf{gpt1} & \textbf{gpt2} & \textbf{gpt3} & \textbf{gpt4} & \textbf{ds1} & \textbf{ds2} & \textbf{ds3} & \textbf{ds4} \\
\midrule
\multirow{12}{*}{\rotatebox[origin=c]{90}{\textbf{Training set}} } & \textbf{qw1} & \heat{0.68} & \heat{0.50} & \heat{0.52} & \heat{0.50} & \heat{0.51} & \heat{0.51} & \heat{0.53} & \heat{0.49} & \heat{0.51} & \heat{0.50} & \heat{0.48} & \heat{0.48} \\
& \textbf{qw2} & \heat{0.53} & \heat{0.62} & \heat{0.54} & \heat{0.60} & \heat{0.52} & \heat{0.51} & \heat{0.53} & \heat{0.51} & \heat{0.51} & \heat{0.52} & \heat{0.49} & \heat{0.49} \\
& \textbf{qw3}  & \heat{0.52} & \heat{0.53} & \heat{0.55} & \heat{0.54} & \heat{0.51} & \heat{0.51} & \heat{0.53} & \heat{0.49} & \heat{0.49} & \heat{0.49} & \heat{0.48} & \heat{0.47} \\
& \textbf{qw4}  & \heat{0.52} & \heat{0.55} & \heat{0.55} & \heat{0.52} & \heat{0.49} & \heat{0.50} & \heat{0.53} & \heat{0.49} & \heat{0.48} & \heat{0.50} & \heat{0.48} & \heat{0.47} \\
& \textbf{gpt1}  & \heat{0.55} & \heat{0.52} & \heat{0.54} & \heat{0.53} & \heat{0.65} & \heat{0.52} & \heat{0.54} & \heat{0.52} & \heat{0.51} & \heat{0.55} & \heat{0.50} & \heat{0.50} \\
& \textbf{gpt2}  & \heat{0.52} & \heat{0.50} & \heat{0.52} & \heat{0.50} & \heat{0.51} & \heat{0.63} & \heat{0.52} & \heat{0.51} & \heat{0.50} & \heat{0.50} & \heat{0.49} & \heat{0.50} \\
& \textbf{gpt3}  & \heat{0.47} & \heat{0.39} & \heat{0.50} & \heat{0.40} & \heat{0.41} & \heat{0.40} & \heat{0.67} & \heat{0.40} & \heat{0.39} & \heat{0.41} & \heat{0.41} & \heat{0.40} \\
& \textbf{gpt4}  & \heat{0.55} & \heat{0.52} & \heat{0.52} & \heat{0.52} & \heat{0.55} & \heat{0.53} & \heat{0.54} & \heat{0.59} & \heat{0.51} & \heat{0.53} & \heat{0.48} & \heat{0.51} \\
& \textbf{ds1} & \heat{0.51} & \heat{0.48} & \heat{0.49} & \heat{0.48} & \heat{0.51} & \heat{0.50} & \heat{0.53} & \heat{0.49} & \heat{0.50} & \heat{0.50} & \heat{0.48} & \heat{0.46} \\
& \textbf{ds2} & \heat{0.45} & \heat{0.43} & \heat{0.47} & \heat{0.44} & \heat{0.43} & \heat{0.43} & \heat{0.51} & \heat{0.43} & \heat{0.43} & \heat{0.45} & \heat{0.42} & \heat{0.42} \\
& \textbf{ds3} & \heat{0.51} & \heat{0.47} & \heat{0.52} & \heat{0.46} & \heat{0.46} & \heat{0.49} & \heat{0.55} & \heat{0.47} & \heat{0.45} & \heat{0.46} & \heat{0.47} & \heat{0.45} \\
& \textbf{ds4} & \heat{0.53} & \heat{0.51} & \heat{0.53} & \heat{0.51} & \heat{0.52} & \heat{0.51} & \heat{0.55} & \heat{0.50} & \heat{0.51} & \heat{0.51} & \heat{0.49} & \heat{0.52} \\
\bottomrule
\end{tabular}
\end{table}

\subsection{Experimental methodology}\label{appDetailsExperimentalMethodology}

\paragraph{Evolutionary approach: standard runs}
Before sampling the KG itself, we first randomly choose the number of entities from $\{5,...,8\}$ and the ratio of edges to entities from $[2.5,3]$. We then use the considered sampler to find a KG with these constraints. 
For the evolutionary search process, we start with an ET which has been trained on 6000 examples. For our main experiments with NoRA, we sample these examples equally from the training, test-D, test-OPEC and test-BL splits. For subsequent rounds, an ET is trained on 6000 examples which are sampled equally from: (i) the training data from the previous round, (ii) the best priority function from the latest round, (iii) test-D, (iv) test-OPEC and (v) test-BL. Further details on the training splits are provided in Section \ref{sec:data-mix}.
We use 10 islands, and for each island $N_{\text{iter}}=20$ attempts are made in each cycle to improve the priority function. Each round consists of four cycles. We set $N_{\text{evalgraphs}}=8$. In each step of the sampling process, $N_{\text{cand}}=20$ candidate edges are scored, with the best-scoring edge being added to the graph.

\paragraph{Evolutionary approach: \textsc{SuperET}}
When applying the evolutionary approach to the \textsc{SuperET}, we make some changes to how the ET is trained after each round, to ensure that the diversity of the initial training data is preserved. In particular, after each round of this process, a new ET is trained on the 16900 original training examples, together with 10400 adversarial examples obtained from the best priority function from the latest round, and 3900 examples obtained from previous rounds (respectively forming 33\% and 12.5\% of the training data source distribution).

\paragraph{LLM service implementation} We have evaluated three LLMs for generating priority functions: deepseek-coder-33b-instruct\footnote{\url{https://huggingface.co/deepseek-ai/deepseek-coder-33b-instruct}}, qwen3-coder-next\footnote{url{https://huggingface.co/Qwen/Qwen3-Coder-Next}}, and gpt-oss-120b\footnote{\url{https://huggingface.co/openai/gpt-oss-120b}}. 
All three models were downloaded from HuggingFace\footnote{\url{https://huggingface.co/}}. In our experiments, the LLMs are served on a single GPU or on multiple GPUs using the vLLM library\footnote{\url{https://vllm.ai/}}. They receive OpenAI API requests sent from the FunSearch algorithm workers, during the evolutionary cycles steps, at each round.  
Since we found in our experiments that the results are largely insensitive to the chosen model, we believe that experiments with larger LLMs would not justify the cost. In particular, the size of the current models allows us to keep the computation time reasonable (about 40 hours per 5-round run in the main experimental setting of this work), while setting the model context length to a size that leaves enough space for the generation of long priority functions. In particular, deepseek-coder-33b-instruct has a 16K maximum model length. For qwen3-coder-next and gpt-oss-120b, we set the maximal context length to 32K because they tend to generate longer outputs.

\subsection{Training data mix strategy}
\label{sec:data-mix}

We provide further details on the training data mix strategy that was used for the evolutionary runs, including the Auto-research experiment. Note that, as explained in Section \ref{appDetailsExperimentalMethodology}, the \textsc{SuperET} was trained with a different strategy.

At every round $i$, three types of data contribute to the
training set.
\begin{enumerate}
    \item \textbf{Priority-generated queries}
The priority function discovered in round $i$ is used to
generate a fresh set of NoRA worlds with 5--8 entities.
For each generated world, $C$ candidate facts are scored by the priority function to serve as queries.
In our experiments we generate $60$ stories per round, yielding roughly $60 \times C$ queries.  

\item \textbf{Historical data}
The training data from previous rounds (for $i>0$) is retained, so the model still has access to previously discovered hard problem instances.
\item \textbf{External datasets}
A fixed collection of external training sets are obtained from the original NoRA dataset, in particular the train split and the 
\texttt{test-BL}, \texttt{test-D} and \texttt{test-OPEC} splits. 
\end{enumerate}

The training data for the ET in each round is capped to 6000 queries. 
To prevent any single source from dominating the training signal, these queries are sampled uniformly from each of the different sources. All sampling uses a fixed random seed for reproducibility. For the first round, only the external datasets are used for training.

\subsection{Auto-Research Agent Experiment}
\label{sec:auto-research}

To test whether the difficulty signals discovered by FunSearch are
artefacts of its particular evolutionary search procedure, we ran an
independent replication using a single LLM agent (Claude) as an
autonomous researcher.  The agent iterates over priority functions one
at a time removing every structural bias introduced by FunSearch.

%% -------------------------------------------------------------------
\subsubsection{Setup}
\label{sec:ar-setup}
The agent operates in a sequential outer loop over \emph{rounds}.
At round~$r$ it:
\begin{enumerate}
  \item reads the evaluation artifacts from round~$r{-}1$
        (accuracy table, sample evaluation instances, the previous priority function (whilst all the rest are also in its context));
  \item proposes a single new priority function;
  \item generates training and evaluation data using the priority function;
  \item trains a fresh Edge Transformer on the merged training set
        and evaluates it.
\end{enumerate}
The same data mix, merge procedure, and ET hyperparameters described previously are used.
The initial model is~\textsc{ET$_0$}, pretrained on the baseline data with no adversarial instances.

\subsubsection{Auto-research recovered Difficulty Signals}
\label{sec:ar-signals}

The priority functions discovered by the agent converged on four
structural signals, each independently rediscovering axes found by the
FunSearch runs:

\begin{enumerate}
  \item \textbf{Maternal/paternal predicates.}
    Facts triggering gender-qualified extended-family relations
    (e.g.\ \texttt{maternal\_grandmother\_of},
    \texttt{paternal\_uncle\_of}) receive the highest weight
    ($12\times$ in the final function).  These require multi-step
    reasoning that chains a gender constraint through an intermediate
    relationship.

  \item \textbf{Multi-hop relations.} 
    Grandparent, aunt/uncle, nibling, and in-law predicates, all
    requiring ${\geq}2$ inference steps, are
    weighted~$5\times$.

  \item \textbf{Entity-pair relationship density.} 
    The agent discovered that entity pairs connected by ${\geq}3$
    distinct relations produce instances the ET struggles with.
    This targets multi-label prediction difficulty rather than
    inference depth.

  \item \textbf{Self-referential fact suppression }
    Unary property facts (same source and target entity, e.g.\
    \texttt{is\_male(3,3)}) are penalised, forcing the generator to
    select relational facts that create richer graph structure.
\end{enumerate}

These signals align with those found by FunSearch,
despite the agent using no evolutionary operators, no population
diversity, and a single sequential pass.

\subsection{Training the Edge Transformer}
The training recipe is identical for the ET across all scenarios and is outlined in the following sections. 
\paragraph{Multi-label binary cross entropy}
We use a multi-label version of the Binary Cross Entropy (BCE) loss for the multi-label classification setting for all problems. The logits for each class are transformed using a sigmoid function and then the problem is treated as a binary classification problem with a multi-hot target binary vector.
\begin{align*}\label{eqbce}
\mathcal{L}_{\text{BCE}} = \sum_{i \in \mathcal{D},j \in \mathcal{R}} \text{CE}(\sigma(x_{ij}), y_{ij})
\end{align*}
where $i$ is the sample index, $j$ is the relation index, $x_{ij}$ is the predicted logit and the $y_{ij}$ is the one-hot target class label.

\paragraph{Initialization and compute}
All trainable parameters for the models are uniformly initialized. All baseline results that were obtained by us were hyperparameter-tuned using grid search, as detailed below. 
All experiments were conducted using RTX 4090, GH200, and A6000 GPUs. A single training cycle (multipe times per evolutionary round) of the ET can be conducted within a few hours on a single GPU. This includes training and testing a single model on any evaluation split.

\paragraph{Hyperparameter settings}
We use the Adam optimizer~\citep{kingma2017adam}. All the models were hyperparameter tuned using an economical grid search over key parameters. For ET, a grid search was performed over the number of attention heads, hidden dimension size, the number of message passing rounds, and dropout rate. All the optimal hyperparameters are available in the companion code with the manuscript.    

\paragraph{Training Protocol}

Each round trains a fresh Edge Transformer from scratch on
$\mathcal{D}_{\mathrm{train}}$.  The data is split $80/20$ into
training and validation partitions via a random split.  We use the
Adam optimiser with learning rate $10^{-3}$, batch size~$32$, and
train for $100$ epochs with gradient clipping.  Validation accuracy is
checked every $10$ epochs.

\section{Additional experimental analysis}

\subsection{Evaluation on Iron Coast}\label{secEvaluationIronCoast}
Table \ref{tabCrossEvaluationSamplersAtlas} repeats this analysis for the Iron Coast domain. For this experiment, we initialized the evolutionary runs with an ET that was trained on examples generated by the Cl-motif sampler. We compare three other generic Claude-generated samplers with the priority functions that were obtained in three independent runs of the evolutionary method. The results clearly shows that challenging queries can be constructed for the Iron Coast domain, which demonstrates the feasibility of a fully automatic evaluation setup, where the world rules, constraints and samplers are all LLM-generated. The priority functions from the three evolutionary runs appear again similar (with models performing similarity on Evo1--Evo3, regardless of which evolutionary priority function was used for training). The three generic Claude-generated samplers all manage to generate challenging problems.

\subsection{Comparison of different LLMs}\label{secEvaluationLLMComparison}
In Table \ref{tabCrossEvaluationSamplersLLMcomparison}, we present a cross-evaluation of the evolutionary priority functions obtained with three different LLMs, where we completed four independent runs for each LLM. The results show that quality of the priority function remains stable across the three models, with the largest model (gpt-oss-120b) performing on par with a much smaller model (deepseek-coder-33b).

\subsection{Analysis of problem difficulty}\label{secDetailsProblemDifficulty}
Given a trained ET and a set of queries, we train a classifier to predict whether a given query will be answered correctly by the ET. The features used for classification are the inference depth, OPEC and BL of the query. As the classifier, we use a Gradient Boosting Classifier. We used the scikit-learn implementation (sklearn.ensemble.HistGradientBoostingClassifier) with the following hyperparameters: max\_depth = 3, n\_estimators = 100 and learning\_rate = 0.1. Each fitted tree model is trained for 100 boosting iterations using cross-entropy.

For the analysis of problem difficulty, all ET models are first calibrated for 5 epochs on 6000 examples sampled from test-D, test-OPEC and test-BL. This calibration step is intensionally limited, so the unique characteristics of the models are maintained. The aim of this calibration step is to ensure that all models have had some exposure to problems with higher reasoning depth, OPEC and BL values, as our focus is on exploring what forms of difficulty remain beyond these three difficulty metrics. 

\paragraph{Impact of OPEEC}
In Figure \ref{figDifficultyAnalysisEntailedOPEC} in the main paper, we showed the percentage reduction in the \emph{Unexplained} score, after incorporating the number of entailed off-path edges (OPEEC) as a fourth difficulty metric, in addition to inference depth, OPEC and BL. In particular, this percentage reduction is defined as
$$
\textit{Reduction} = 100 \cdot \frac{\textit{Unexplained}_{\textit{base}} - \textit{Unexplained}_{\textit{aug}}}{\textit{Unexplained}_{\textit{base}}}
$$
where $\textit{Unexplained}_{\textit{base}}$ is the \textit{Unexplained} score, as defined by \eqref{eqUnexplained}, for the classifier that only incorporates inference depth, OPEC and BL; $\textit{Unexplained}_{\textit{aug}}$ is the corresponding value for the augmented classifier, which additionally incorporates the number of entailed off-path edges. We report the median value of \textit{Reduction} that was found across 100 paired bootstrap samples (where each bootstrap sample has the same size as the ET training set for that model). To test for statistical significance, we used a one-sided empirical paired bootstrap test, where the null hypothesis is that $\textit{Reduction}\leq 0$. The $p$-value is computed as 
$$
p = \frac{1 + \text{number of bootstrap samples with $\textit{Reduction}\leq 0$}}{1+100}
$$
Figure \ref{figDifficultyAnalysisEntailedOPEC} shows the value of $\textit{Reduction}$ for some of the models. As can be seen, the values are positive for all tested models.  The fact that the reduction is positive is statistically significant for all models ($p<0.05$).

%********************
\subsection{Analysis of the evolved priority functions}\label{sec:ast}

We analyze the best evolved priority functions retained at the end of each evolutionary round using abstract syntax tree (AST) tokens. 
For each priority function, we parse the
source with the standard Python \texttt{ast} module and extract four kinds of
pseudo-tokens: \emph{calls}, \emph{attribute accesses},  \emph{operators} and \emph{imports}.
%\emph{calls} (e.g. \texttt{call:foo}, \texttt{calldot:obj.foo}),
%\emph{attribute accesses} (\texttt{attr:bar}), \emph{operators}
%(\texttt{op\_and}, \texttt{op\_in}, \texttt{op\_eq},
%\texttt{op\_gte} etc.), and \emph{imports}
%(\texttt{import:networkx}, etc.). 
The same parser also records structural metrics: the
number of AST nodes, the maximum nesting depth, the number of \texttt{Call} nodes, the
number of \texttt{Compare} nodes, the number of branching nodes (\texttt{If},
\texttt{For}, \texttt{While}, \texttt{Try}, etc.), and the number of distinct tokens.
Across Evo1--Evo4, this yields 20 priority functions and 158 distinct raw tokens,
giving a $20\times158$ token-count matrix.

\begin{figure}
  \centering
  \begin{subfigure}{\linewidth}
    \centering
    \includegraphics[width=0.8\linewidth]{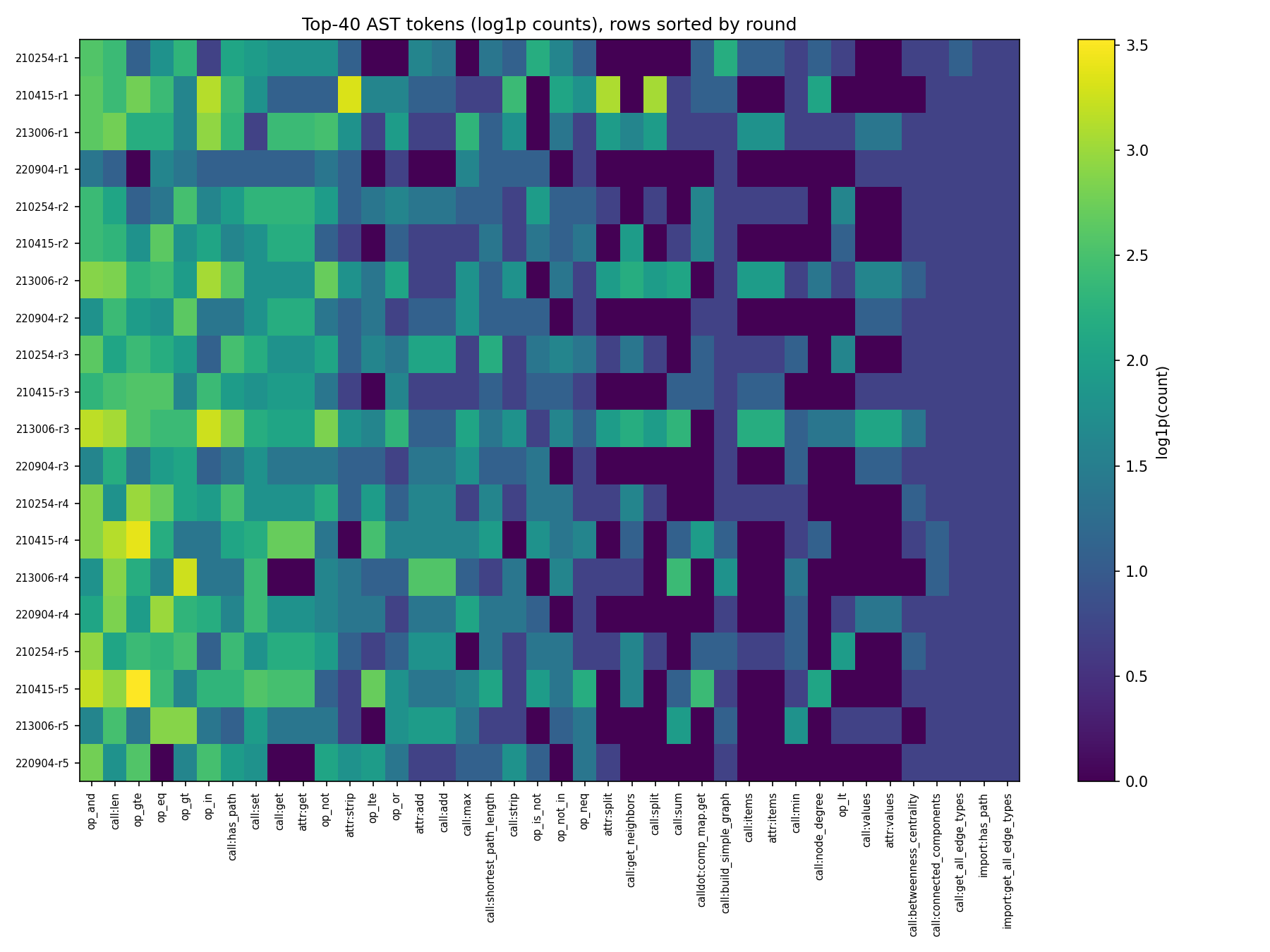}
    \caption{}
    \label{fig:ast:heatmap}
  \end{subfigure}

  \vspace{0.5em}

  \begin{subfigure}{\linewidth}
    \centering
    \includegraphics[width=0.6\linewidth]{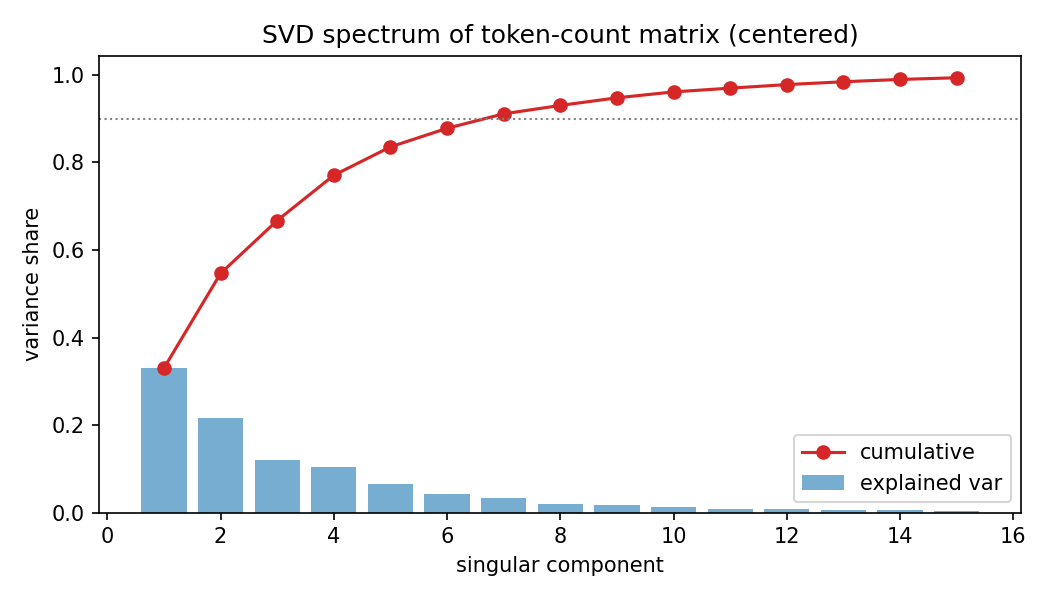}
    \caption{}
    \label{fig:ast:svd-spectrum}
  \end{subfigure}

  \vspace{0.5em}

  \begin{subfigure}{\linewidth}
    \centering
    \includegraphics[
      width=\linewidth,
      trim={0 0 0 2.3cm},
      clip
    ]{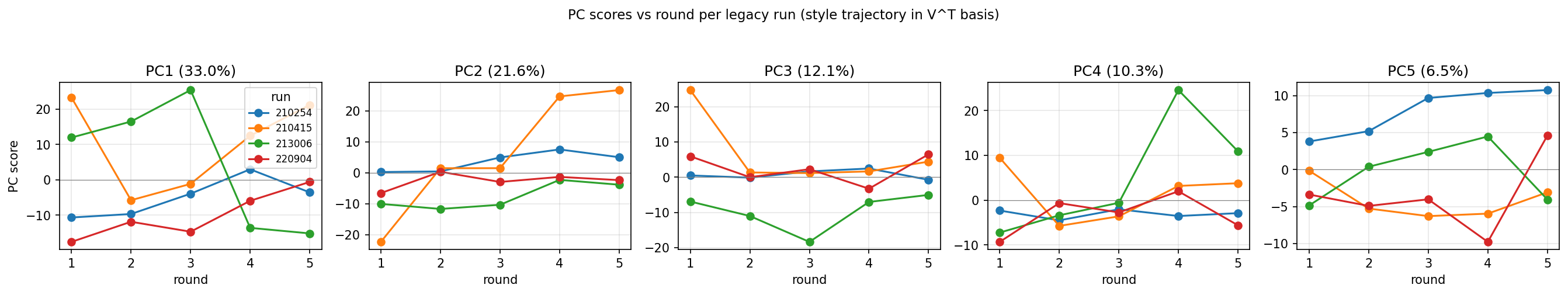}
    \caption{}
    \label{fig:ast:svd-traj}
  \end{subfigure}
  \caption{(a) AST token heatmap: $\log(1{+}\textrm{count})$ for the top-$40$
  tokens across the $20$ priority functions. (b) SVD spectrum of the
  column-centered token-count matrix: per-component variance share (bars) and
  cumulative share (line). The horizontal line marks $90\%$ explained variance.
  (c) Trajectory of each Evolutionary Search run (Evo1--Evo4) in the PC$_1$--PC$_5$
  space across rounds.}
  \label{fig:ast:combined}
\end{figure}

Figure~\ref{fig:ast:heatmap} shows the $\log(1+\textrm{count})$
abundance of the top-$40$ most frequent tokens, with rows ordered by
round and grouped by run. The dominant columns are logical and relational
operators; the graph related tokens (in the right hand side of the heat-map) are sparse,
%(\texttt{op\_and}, \texttt{op\_in}, \texttt{op\_gte},
%\texttt{op\_eq}) together with \texttt{call:len},
%\texttt{call:has\_path} and \texttt{call:get}. The right-hand block of
%the matrix (e.g.\ \texttt{call:get\_all\_edge\_types},
%\texttt{call:betweenness\_centrality},
%\texttt{call:connected\_components})
indicating occasional spikes in
graph-related queries by individual runs. Centering the token matrix column-wise and
taking its SVD gives us a low-dimensional
structure in token usage (Figure~\ref{fig:ast:svd-spectrum}): the
first two singular components together explain roughly $55\%$ of the
variance, and the first seven explain at least $90\%$. Figure~\ref{fig:ast:svd-traj}
then projects the evolution of priority functions onto the first five principal components and traces the trajectory
of each run across rounds.

  \begin{figure}[t]
    \centering
    \includegraphics[width=\linewidth]{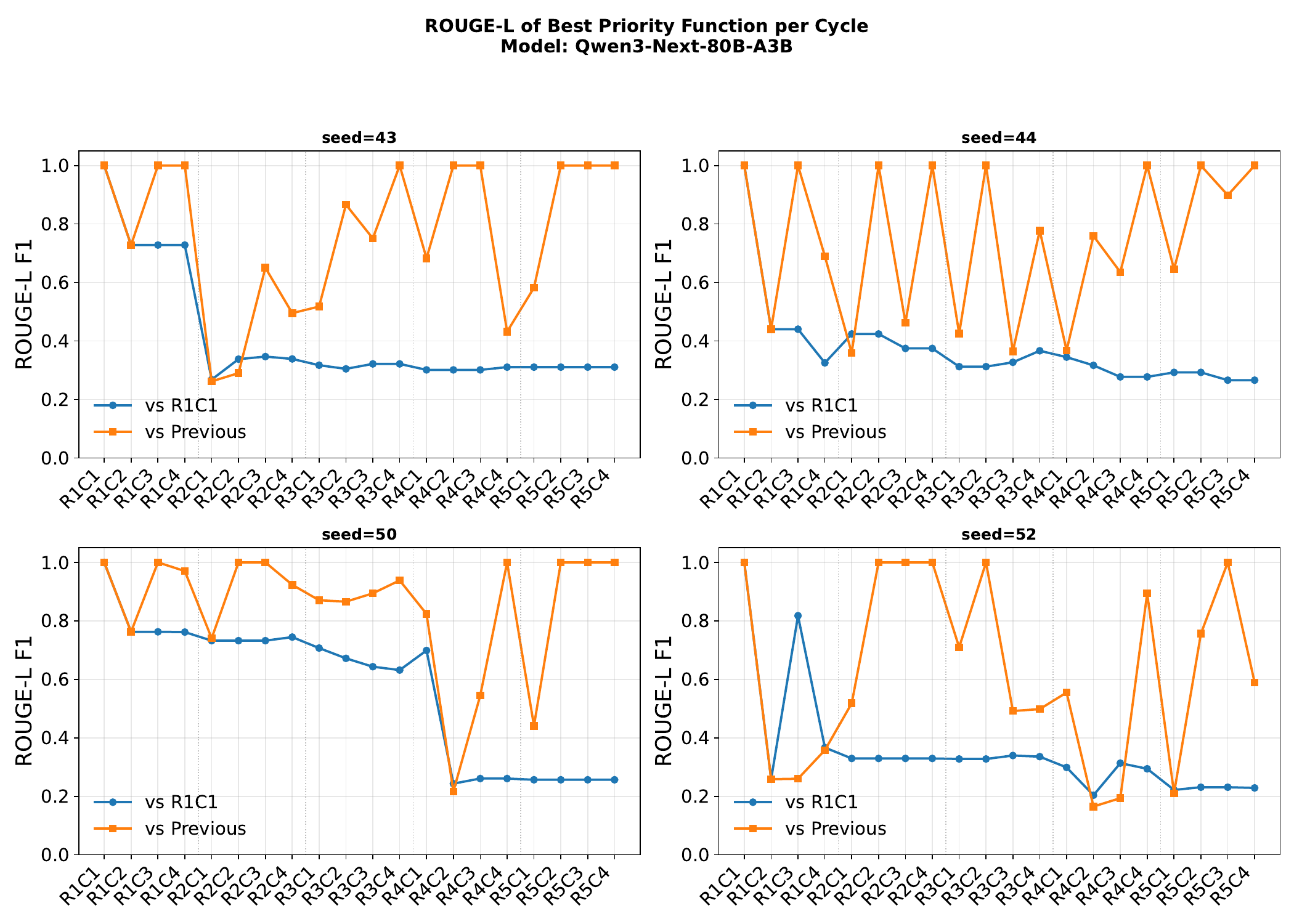}
    \caption{Analysis of the syntactic similarity between the evolved priority functions and the initial one (blue) and the priority function from the previous cycle (orange), measured in terms of ROUGE-L.}
    \label{fig:rouge-l}
  \end{figure}
  \begin{figure}[t]
    \centering
    \includegraphics[width=\linewidth]{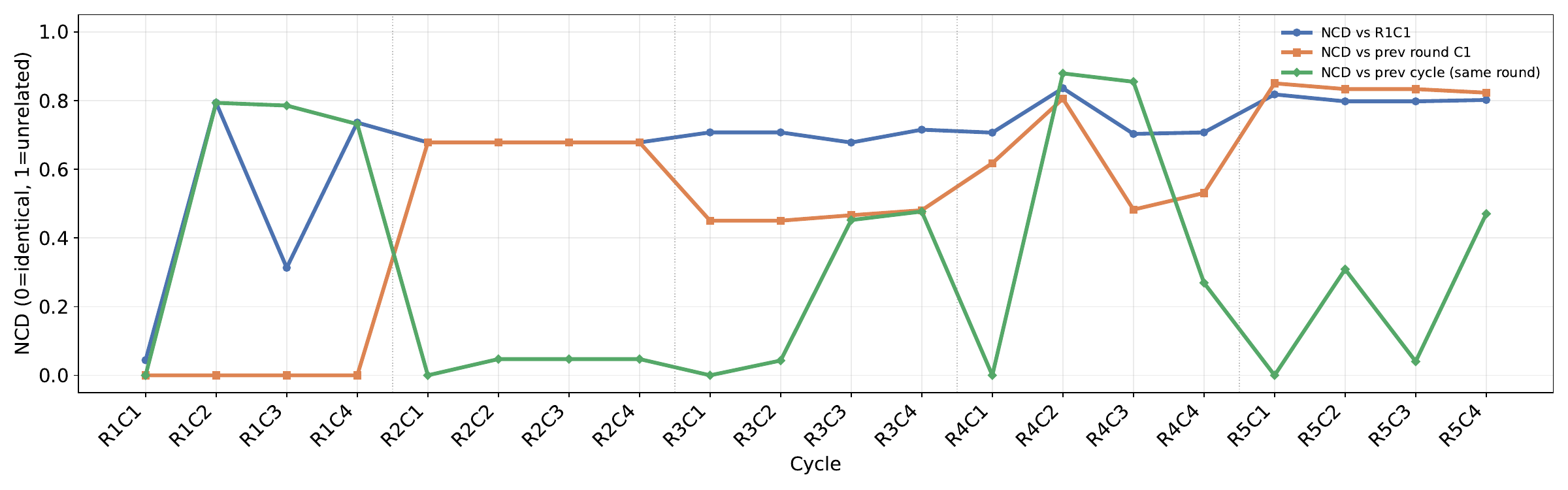}
    \caption{Analysis of the Normalized Compression Distance of the learned priority functions at the current cycle with respect various reference point priority functions.}
    \label{fig:ncd}
  \end{figure}
  \begin{figure}[h]
    \centering
    \includegraphics[width=\linewidth]{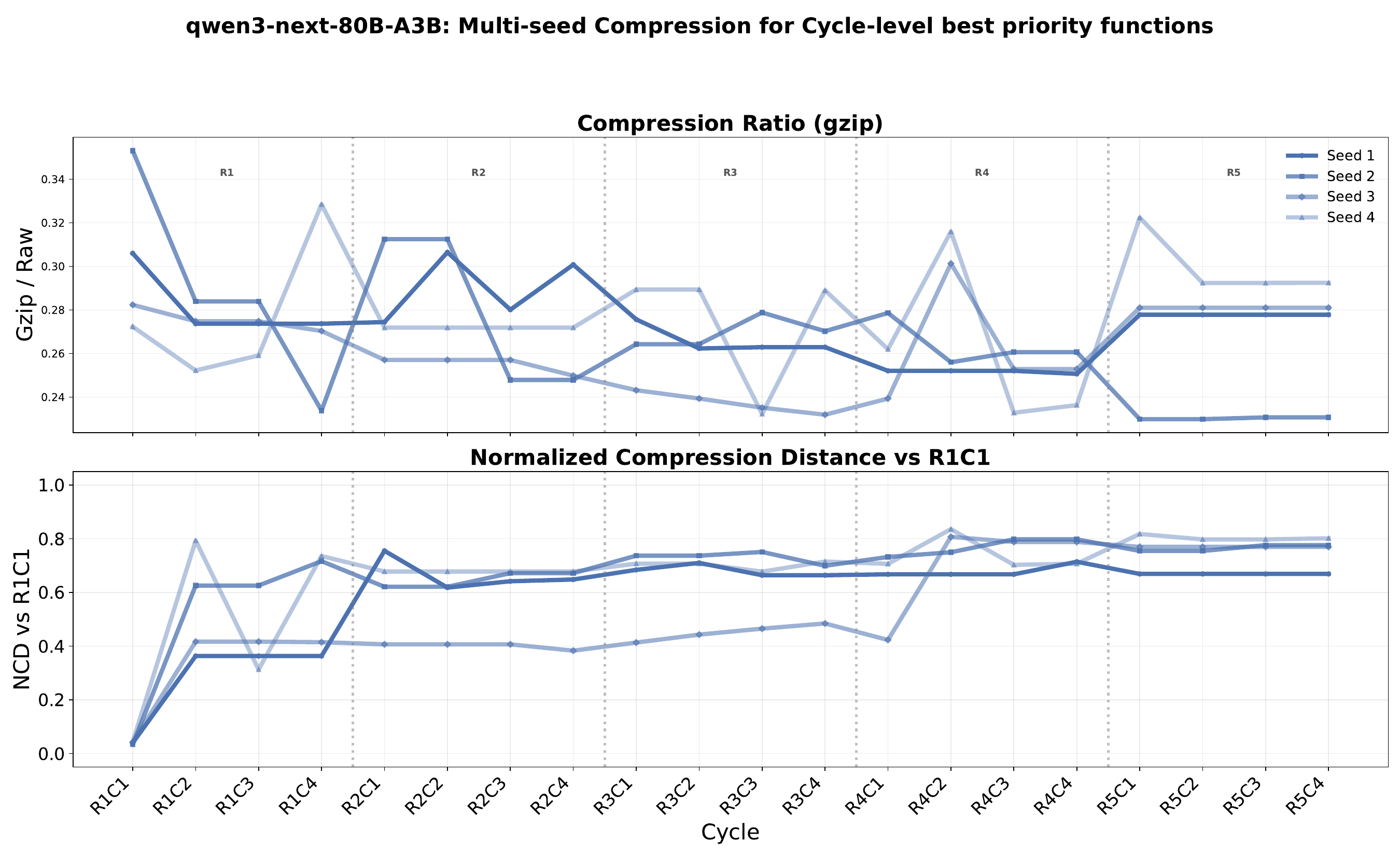}
    \caption{Analysis of the Normalized Compression Distance (NCD) and compression ratio of the learned priority functions at the current cycle with respect to the best priority function at the end of round 1 for multiple seeds. This is a heuristic proxy for Kolmogorov complexity and again confirms what ROUGE-L evolution suggests that syntactically, the priority functions are changing less with respect to their states at the end of round 1.}
    \label{fig:ncd-seed}
  \end{figure}

Figure \ref{fig:rouge-l} shows an analysis of how much the priority function changes throughout the evolutionary process. For this analysis, we compute the syntactic similarity between the best priority function obtained at the end of each cycle and (i) the priority function obtained in the first cycle and (ii) the priority function obtained in the previous cycle. Syntactic similarity is measured in terms of ROUGE-L. We can see that large changes are infrequent and may occur in different phases of the evolutionary process.

Figure \ref{fig:ncd} shows how the complexity of the priority function evolves, estimated by its normalized compression distance $\mathrm{NCD}(x,y) = \frac{C(xy) - \min\{C(x), C(y)\}}{\max\{C(x), C(y)\}}$  where $C(\cdot)$ is the compressed length and $xy$ denotes concatenation.

\subsection{Analysis of graphs generated by the evolution process}
\begin{figure}[t]
    \centering
    \includegraphics[width=\linewidth]{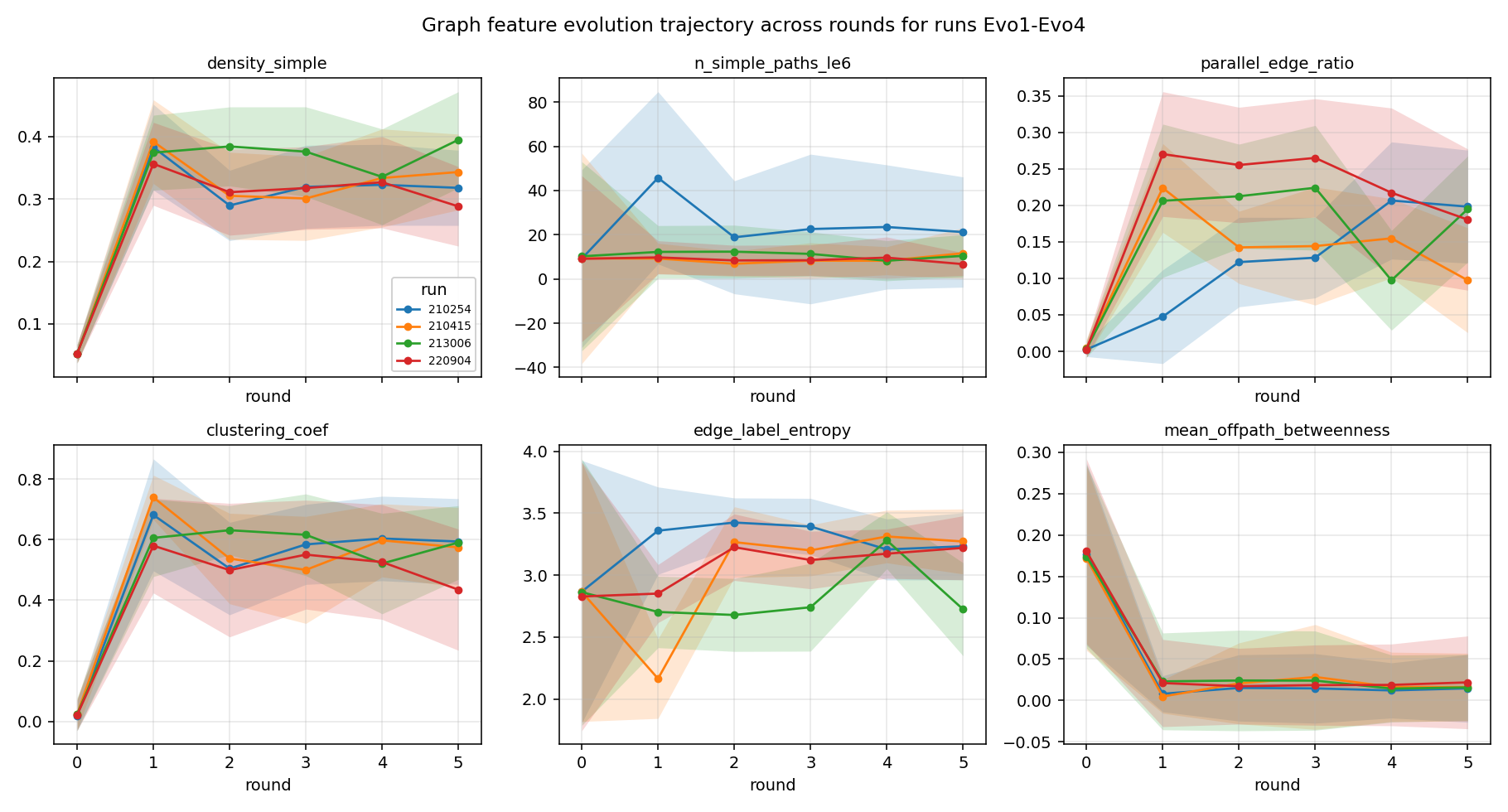}
    \caption{\textsc{Evolutionary search}: Evolution of mean graph features, across 5 rounds, for Funsearch runs Evo1-Evo4.}
    \label{fig:evo:trajectories}
  \end{figure}

\begin{figure}[t]
    \centering
    \includegraphics[width=\linewidth]{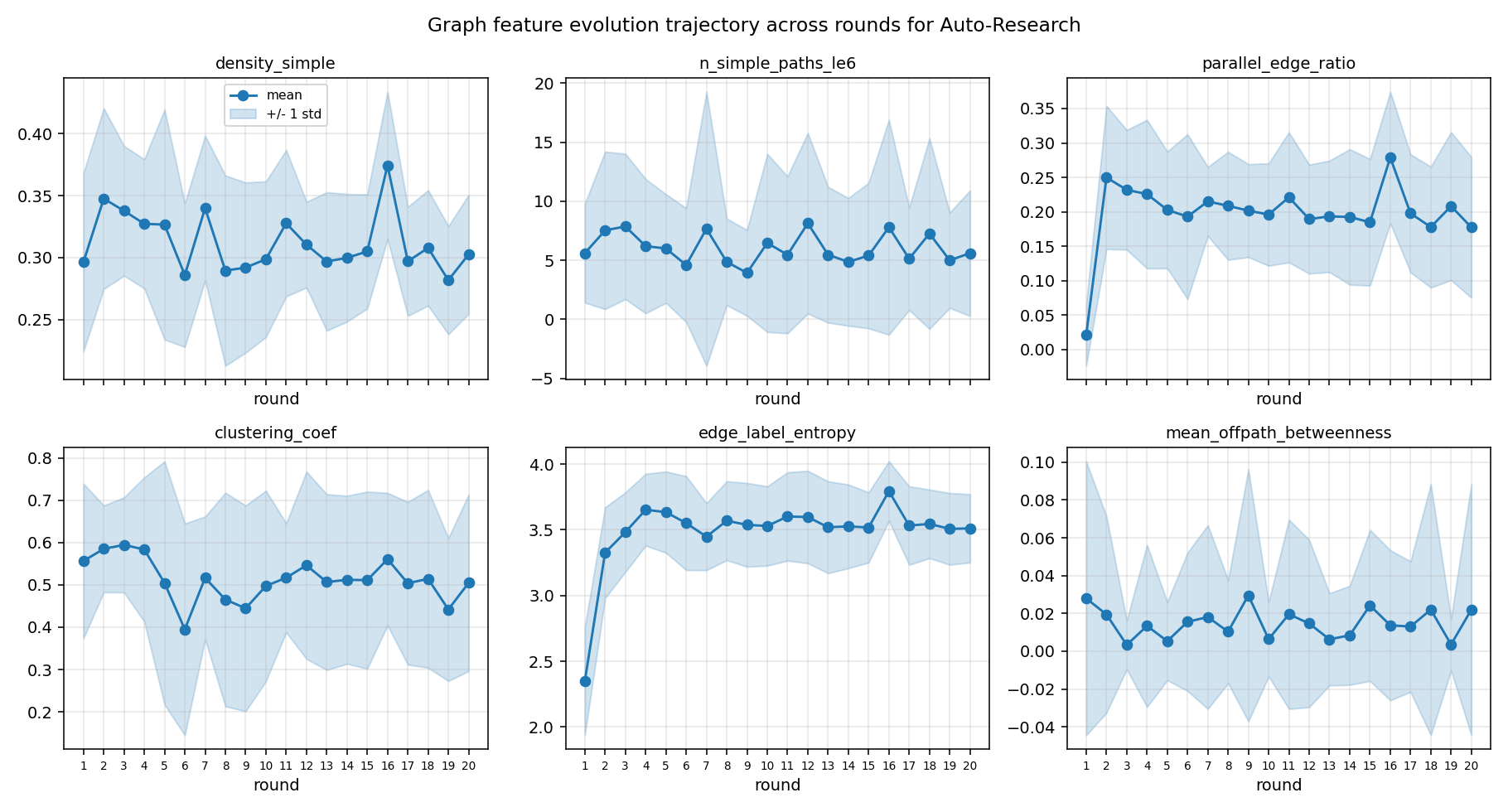}
    \caption{\textsc{Auto-research}: Evolution of mean graph features, across 20 rounds.}
    \label{fig:ar:trajectories}
  \end{figure}

\begin{figure}[t]
    \centering
    \includegraphics[width=\linewidth]{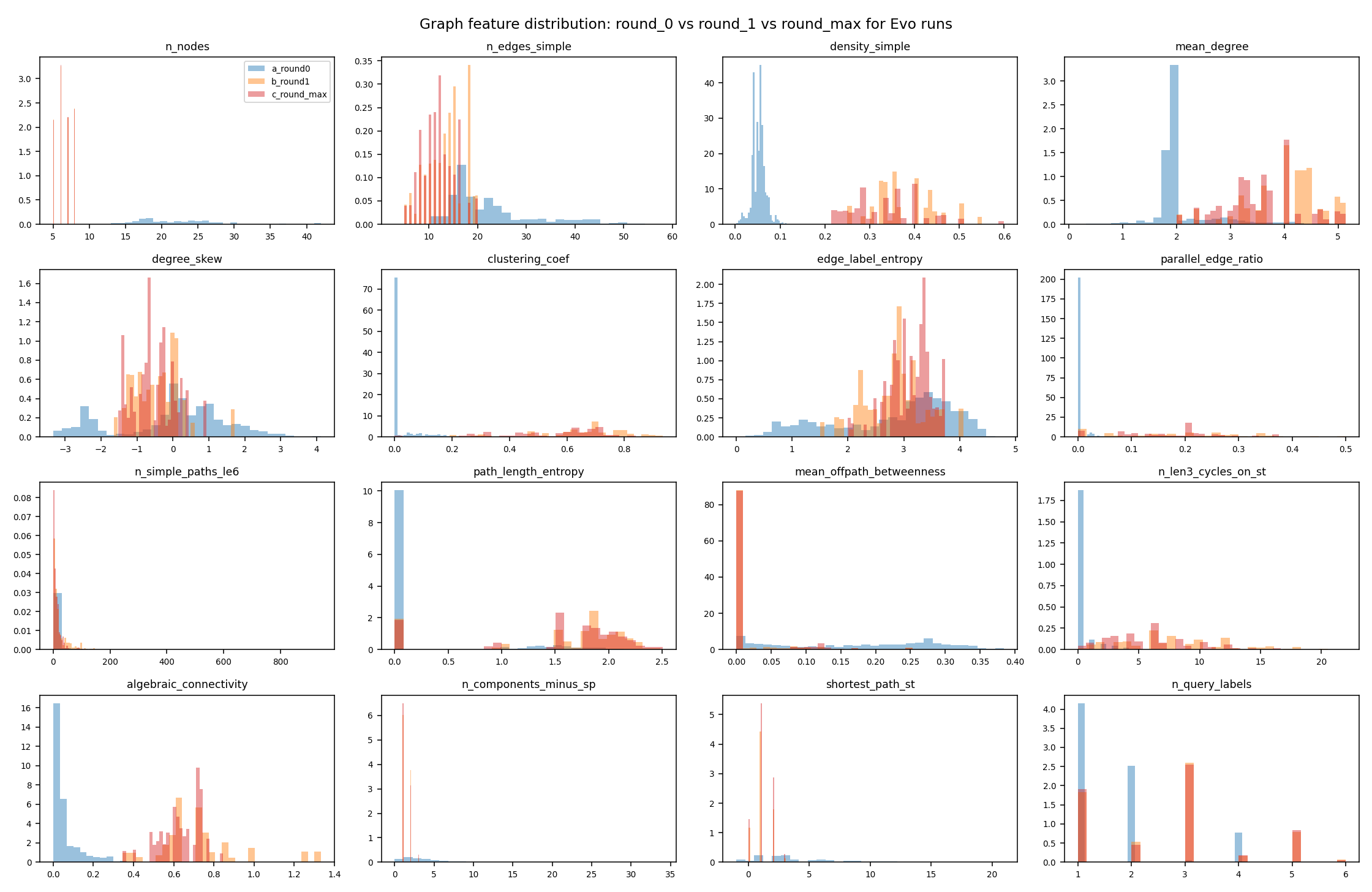}
    \caption{\textsc{Evolutionary search}: histograms of selected graph
    features for round 0, round 1 and round 5.}
    \label{fig:evo:hist}
  \end{figure}

\begin{figure}
    \centering
    \includegraphics[width=\linewidth]{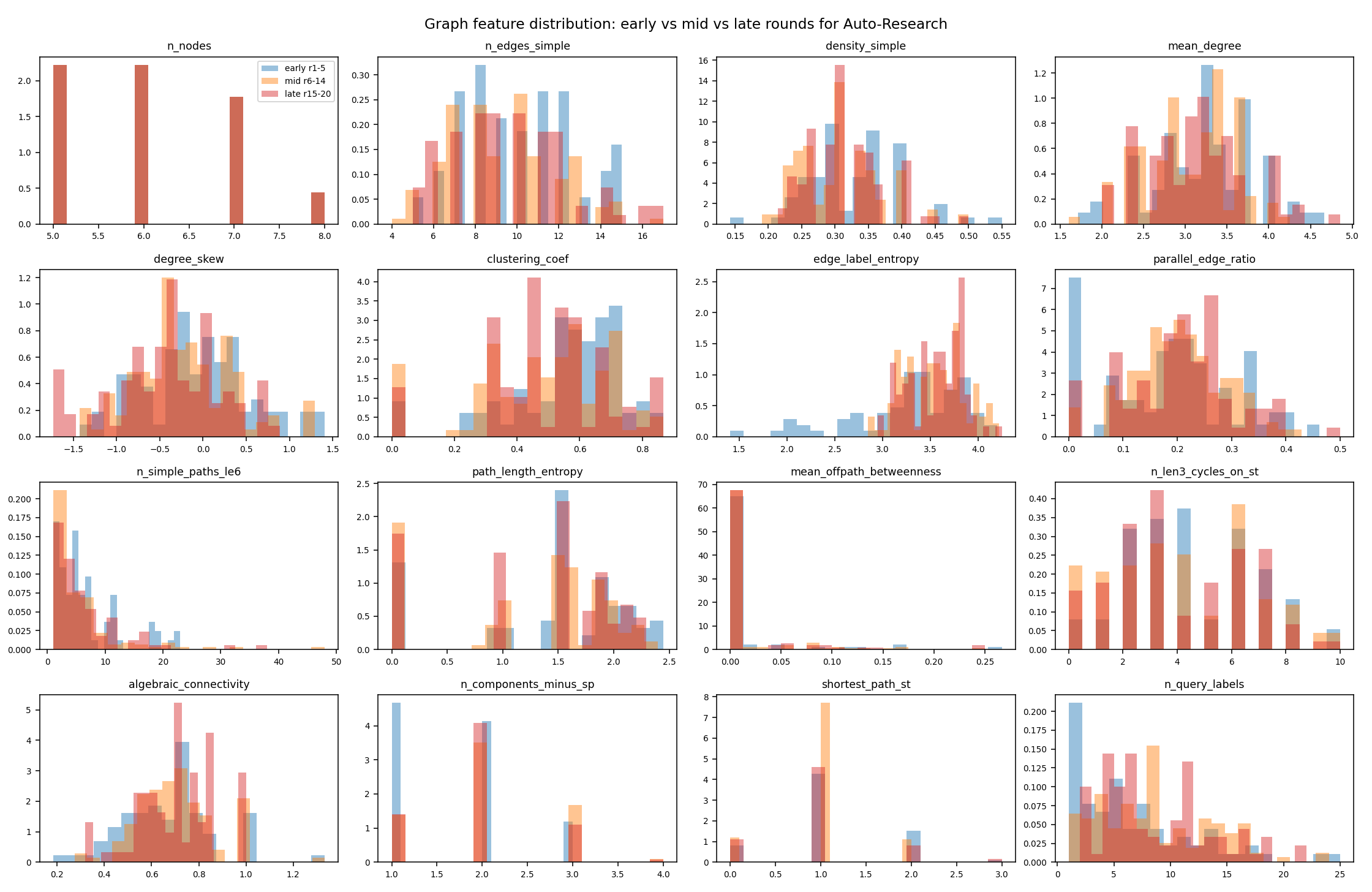}
    \caption{\textsc{Auto-research}: histograms of selected features for
    early ($r$=1--5), mid ($r$=6--14) and late ($r$=15--20) phases.}
    \label{fig:ar:hist}
  \end{figure}

In this section, we perform an exploratory analysis of the knowledge graphs produced by the priority functions from the evolutionary runs (Evo1–Evo4) and the Claude auto-research protocol. Specifically, given that the graph size remains fixed, we investigate how density, path lengths, and the number of parallel edges evolve across rounds, making reasoning a difficult task for the edge transformer. We test 20 graph connectivity features (derived using \texttt{NetworkX}), which can be broadly classified into global structure, source-target reasoning paths, spectral/topological connectivity, and number of possible query labels. We note that evolutionary process has the highest jump (increase/decrease) in most graph features from round zero to round one (see Figures \ref{fig:evo:trajectories} and \ref{fig:ar:trajectories}) for both sampling methods. The feature edge label entropy (diversity of relational labels between two nodes), parallel edge ratio (fraction of fact-edges that are redundant in the sense that some other relation already connects the same node pair), and the number of potential query labels shows consistent increase from the baseline (Figures \ref{fig:evo:hist} and \ref{fig:ar:hist}).

  \hfill

  \vspace{0.75em}

  Graph feature trajectories and histogram summaries for the two
  sampler families are shown in Figures \ref{fig:evo:trajectories}, \ref{fig:ar:trajectories}, \ref{fig:evo:hist},  \ref{fig:ar:hist}. Both sampling methods show increases in the number of parallel edges and edge-label entropy, increasing ambiguity for the edge transformer.

\subsection{Example priority functions}
In this section we present the best priority functions generated by all the approaches used in this paper. The FunSearch function is from a Qwen3-80B-A3B model run for 5 rounds with 4 cycles. Note that there are general and world specific features in all the functions. The Auto-Research functions are much more compact and compositional in nature. For instance, round 7 has mostly general features whereas round 20 is more world specific but also contain some features from round 7. 
\begin{figure*}[h!]
\lstset{
  keywordstyle=\color{blue!70!black}\bfseries,
  commentstyle=\color{green!50!black}\itshape,
  stringstyle=\color{purple!70!black},
}
\begin{tcolorbox}[colback=teal!5!white,colframe=teal!75!black,title=\textsc{Round 20 Auto-research (best priority function)}, width=\linewidth, nobeforeafter]
\footnotesize
\begin{lstlisting}[language=Python,basicstyle=\ttfamily\footnotesize,breaklines=true,showstringspaces=false]
def priority(cand_fact: str, definite_rules_program: str, entailed_facts: str, facts_program: str) -> float:
    if not entailed_facts.strip():
        return -10.0

    cand_rel = cand_fact.strip().rstrip('.').split('(')[0] if '(' in cand_fact else ''
    if cand_rel.startswith('not_'):
        return -100.0

    # Expanded trivial set: spatial + attributes + non-family social
    trivial = {'living_in', 'living_in_same_place', 'is_person', 'is_place',
               'is_male', 'is_female', 'is_underage', 'no_brothers', 'no_sisters',
               'no_sons', 'no_daughters', 'colleague_of', 'school_mates_with'}

    multi_hop = {'grandmother_of', 'grandfather_of', 'granddaughter_of', 'grandson_of',
                 'maternal_grandmother_of', 'maternal_grandfather_of',
                 'paternal_grandmother_of', 'paternal_grandfather_of',
                 'aunt_of', 'uncle_of', 'niece_of', 'nephew_of',
                 'maternal_aunt_or_uncle_of', 'paternal_aunt_or_uncle_of',
                 'maternal_uncle_of', 'paternal_uncle_of',
                 'great_grandparent_of', 'great_grandchild_of',
                 'sister_in_law_of', 'brother_in_law_of',
                 'father_in_law_of', 'mother_in_law_of',
                 'son_in_law_of', 'daughter_in_law_of'}

    lines = entailed_facts.strip().split('\n')

    existing_rels = set()
    for line in facts_program.strip().split('\n'):
        line = line.strip().rstrip('.')
        if '(' in line:
            existing_rels.add(line.split('(')[0])

    score = 0.0
    entailed_rels = set()
    multi_hop_count = 0
    for line in lines:
        line = line.strip().rstrip('.')
        if '(' in line:
            rel = line.split('(')[0]
            if rel.startswith('not_'):
                continue
            entailed_rels.add(rel)
            if rel in trivial:
                score -= 2.0
            elif rel in multi_hop:
                score += 20.0
                multi_hop_count += 1
            else:
                score += 1.0

    novel_rels = entailed_rels - existing_rels - trivial
    score += len(novel_rels) * 10.0
    non_trivial = entailed_rels - trivial
    score += len(non_trivial) * 3.0
    score += multi_hop_count * 8.0

    return score
\end{lstlisting}
\end{tcolorbox}
\end{figure*}

\begin{figure*}[h!]
\lstset{
  keywordstyle=\color{blue!70!black}\bfseries,
  commentstyle=\color{green!50!black}\itshape,
  stringstyle=\color{purple!70!black},
}
\begin{tcolorbox}[colback=teal!5!white,colframe=teal!75!black,title=\textsc{Round 5 Cycle 4 FunSearch (best priority function)}, width=\linewidth, nobeforeafter]
\tiny
\begin{lstlisting}[language=Python,basicstyle=\ttfamily\scriptsize,breaklines=true,showstringspaces=false,columns=fullflexible]
def priority(cand_fact: str, definite_rules_program: str, entailed_facts: str, facts_program: str) -> float:
    """Improved version of `priority` that adds scoring for entailed edges that create multi-hop cycles and penalizes redundant edge types."""
    from Funsearch.GraphWranglingMethods import betweenness_centrality, build_simple_graph, connected_components, count_paths, get_all_edge_types, has_path, num_nodes, shortest_path_length
    import re

    # Parse the candidate edge
    try:
        cand_fact_stripped = cand_fact.strip()
        match = re.match(r'^(\w+)\s*\(\s*(\d+)\s*,\s*(\d+)\s*\)\s*\.\s*$', cand_fact_stripped)
        if not match:
            return 0.0
        relation, source, target = match.groups()
    except Exception:
        return 0.0

    # Parse the entailed facts into edges
    try:
        entailed_lines = [l.strip() for l in entailed_facts.strip().split('\n') if l.strip() and not l.startswith('#')]
    except Exception:
        entailed_lines = []

    if not entailed_lines:
        return 0.0

    # Basic score: number of entailed edges with base multiplier
    score = len(entailed_lines) * 0.35

    # Relation type weights (tuned for fine-grained differentiation)
    if relation in ('parent_of', 'sibling_of', 'spouse_of'):
        score += 1.6
    if relation in ('grandparent_of', 'uncle_of', 'aunt_of', 'nibling_of'):
        score += 0.85
    if relation in ('living_in_same_place', 'school_mates_with', 'colleague_of'):
        score += 0.55

    # Structural analysis
    try:
        G = build_simple_graph(facts_program)
        betweenness = betweenness_centrality(G)
        original_edge_types = get_all_edge_types(G)
        edge_types_seen = set()
        long_range_bonus = 0.0
        bridge_bonus = 0.0
        new_reach_bonus = 0.0
        chain_potential_bonus = 0.0
        indirect_bonus = 0.0
        constraint_bonus = 0.0
        multi_hop_cycle_bonus = 0.0
        redundancy_penalty = 0.0

        n = num_nodes(G)

        # Analyze each entailed edge
        for line in entailed_lines:
            try:
                em = re.match(r'^(\w+)\s*\(\s*(\d+)\s*,\s*(\d+)\s*\)\s*\.\s*$', line.strip())
                if not em:
                    continue
                rel_name, src, tgt = em.groups()
                edge_types_seen.add(rel_name)

                # Enhanced novel reachability scoring
                if not has_path(G, src, tgt):
                    score += 2.3
                    new_reach_bonus += 0.5 * (n / max(n, 4))
\end{lstlisting}
\end{tcolorbox}
\end{figure*}

\begin{figure*}[h!]
\lstset{
  keywordstyle=\color{blue!70!black}\bfseries,
  commentstyle=\color{green!50!black}\itshape,
  stringstyle=\color{purple!70!black},
}
\begin{tcolorbox}[colback=teal!5!white,colframe=teal!75!black,title=..., width=\linewidth, nobeforeafter]
\tiny
\begin{lstlisting}[language=Python,basicstyle=\ttfamily\scriptsize,breaklines=true,showstringspaces=false,columns=fullflexible]
                else:
                    pl = shortest_path_length(G, src, tgt)
                    if pl is not None:
                        if pl >= 6:
                            long_range_bonus += 3.0
                        elif pl >= 5:
                            long_range_bonus += 2.0
                        elif pl >= 4:
                            long_range_bonus += 1.3
                        elif pl >= 3:
                            long_range_bonus += 0.7
                # Bridge detection with stricter thresholds
                if src in betweenness and tgt in betweenness:
                    avg_bc = (betweenness[src] + betweenness[tgt]) / 2
                    if not has_path(G, src, tgt) and avg_bc > 0.14:
                        bridge_bonus += 2.6
                    elif avg_bc > 0.10:
                        bridge_bonus += 0.6

                # Novelty scoring with tighter penalty/bonus balance
                if rel_name in original_edge_types:
                    score -= 0.25
                    redundancy_penalty += 0.1  # Track redundant types for bonus/penalty calculation
                else:
                    score += 0.75

                # Shortcut detection (counting paths up to 5 hops for better sensitivity)
                paths_count = count_paths(G, src, tgt, cutoff=5)
                if paths_count >= 2 and has_path(G, src, tgt):
                    score += 0.6

                # Constraint sensitivity: stronger bonus for distinct nodes
                if src != tgt:
                    constraint_bonus += 0.2

                # Chain potential bonus: slightly increased weight
                chain_relations = {'child_of', 'parent_of', 'grandparent_of', 'sibling_of', 'spouse_of'}
                if rel_name in chain_relations:
                    chain_potential_bonus += 0.28

                # Indirect edge bonus: detect if entailed edge has high betweenness endpoints
                try:
                    if src in betweenness and tgt in betweenness:
                        bc_product = betweenness[src] * betweenness[tgt]
                        if bc_product > 0.006 and not has_path(G, src, tgt):
                            indirect_bonus += 0.35
                except:
                    pass

                # Multi-hop cycle detection: detect cycles with path length >= 2
                if src != tgt and has_path(G, src, tgt):
                    path_len_to_tgt = shortest_path_length(G, src, tgt)
                    if path_len_to_tgt is not None and path_len_to_tgt >= 2:
                        if has_path(G, tgt, src):
                            multi_hop_cycle_bonus += 1.1 * (path_len_to_tgt / max(path_len_to_tgt, 2))

            except Exception:
                continue

        # Add bonuses with optimized weights
        score += bridge_bonus * 1.5
        score += long_range_bonus * 1.1
        score += new_reach_bonus * 1.2
        score += chain_potential_bonus * 1.8
        score += indirect_bonus * 1.4
        score += constraint_bonus * 1.5
        score += multi_hop_cycle_bonus * 1.3  # NEW: emphasize multi-hop cycles more strongly

        # Diversity bonus: slightly stronger nonlinearity and cap the bonus
        score += 1.0 * (len(edge_types_seen) ** 0.82)

        # Redundancy penalty scaling
        score -= redundancy_penalty * 0.4
        
\end{lstlisting}
\end{tcolorbox}
    
\end{figure*}
\begin{figure*}[h!]
\lstset{
  keywordstyle=\color{blue!70!black}\bfseries,
  commentstyle=\color{green!50!black}\itshape,
  stringstyle=\color{purple!70!black},
}
\begin{tcolorbox}[colback=teal!5!white,colframe=teal!75!black,title=..., width=\linewidth, nobeforeafter]
\tiny
\begin{lstlisting}[language=Python,basicstyle=\ttfamily\scriptsize,breaklines=true,showstringspaces=false,columns=fullflexible]
        # Bridge candidate analysis with refined degree criteria
        try:
            src_deg = len(set(G.successors(source)) | set(G.predecessors(source)))
            tgt_deg = len(set(G.successors(target)) | set(G.predecessors(target)))

            # Prefer asymmetric connections (low-degree to high-degree)
            if (src_deg <= 1 and tgt_deg >= 5) or (tgt_deg <= 1 and src_deg >= 5):
                score += 1.5
            elif (src_deg <= 2 and tgt_deg >= 4) or (tgt_deg <= 2 and src_deg >= 4):
                score += 0.85
            elif (src_deg <= 1 and tgt_deg >= 4) or (tgt_deg <= 1 and src_deg >= 4):
                score += 0.75

            # Bonus for connecting components
            components = list(connected_components(G))
            if len(components) > 1:
                src_comp = next((c for c in components if source in c), None)
                tgt_comp = next((c for c in components if target in c), None)
                if src_comp and tgt_comp and src_comp != tgt_comp:
                    score += 2.6
        except Exception:
            pass

    except Exception:
        pass  # Fall back to basic score if graph building fails

    return float(score)
\end{lstlisting}
\end{tcolorbox}
\end{figure*}

\begin{figure*}[h!]
\lstset{
  keywordstyle=\color{blue!70!black}\bfseries,
  commentstyle=\color{green!50!black}\itshape,
  stringstyle=\color{purple!70!black},
}
\begin{tcolorbox}[colback=teal!5!white,colframe=teal!75!black,title=\textsc{Round 7 Auto-Research (best priority function)}, width=\linewidth, nobeforeafter]
\tiny
\begin{lstlisting}[language=Python,basicstyle=\ttfamily\scriptsize,breaklines=true,showstringspaces=false,columns=fullflexible]
def priority(cand_fact: str, definite_rules_program: str, entailed_facts: str, facts_program: str) -> float:
    if not entailed_facts.strip():
        return 0.0

    cand_rel = cand_fact.strip().rstrip('.').split('(')[0] if '(' in cand_fact else ''
    if cand_rel.startswith('not_'):
        return -100.0

    lines = entailed_facts.strip().split('\n')

    # Parse existing entity pairs from the story
    existing_pairs = {}
    for line in facts_program.strip().split('\n'):
        line = line.strip().rstrip('.')
        if '(' in line:
            rel = line.split('(')[0]
            args = line.split('(')[1].rstrip(')').split(',')
            if len(args) == 2:
                pair = (args[0].strip(), args[1].strip())
                existing_pairs.setdefault(pair, set()).add(rel)

    # Score entailed facts: prefer those that ADD relations to existing pairs
    score = 0.0
    entailed_rels = set()
    overlap_count = 0
    for line in lines:
        line = line.strip().rstrip('.')
        if '(' in line:
            rel = line.split('(')[0]
            if rel.startswith('not_'):
                continue
            entailed_rels.add(rel)
            args = line.split('(')[1].rstrip(')').split(',')
            if len(args) == 2:
                pair = (args[0].strip(), args[1].strip())
                if pair in existing_pairs:
                    overlap_count += 1  # adds complexity to existing pairs

    # Prefer: many entailed, many overlapping pairs, diverse relations
    score = float(len(lines)) + overlap_count * 5.0 + len(entailed_rels) * 3.0
    return score
\end{lstlisting}
\end{tcolorbox}
\end{figure*}
%%%%%%%%%%%%%%%%%%%%%%%%%%%%%%%%%%%%%%%%%%%%%%%%%%%%%%%%%%%%
\clearpage
\newpage
\section*{NeurIPS Paper Checklist}

\begin{enumerate}

\item {\bf Claims}
    \item[] Question: Do the main claims made in the abstract and introduction accurately reflect the paper's contributions and scope?
    \item[] Answer: \answerYes{} % Replace by \answerYes{}, \answerNo{}, or \answerNA{}.
    \item[] Justification: The paper and the abstract are aligned in content and scope.

\item {\bf Limitations}
    \item[] Question: Does the paper discuss the limitations of the work performed by the authors?
    \item[] Answer: \answerYes{} % Replace by \answerYes{}, \answerNo{}, or \answerNA{}.
    \item[] Justification: The limitations of this work are discussed in the Conclusions section.

\item {\bf Theory assumptions and proofs}
    \item[] Question: For each theoretical result, does the paper provide the full set of assumptions and a complete (and correct) proof?
    \item[] Answer: \answerNA{} % Replace by \answerYes{}, \answerNo{}, or \answerNA{}.
    \item[] Justification: \answerNA{}

    \item {\bf Experimental result reproducibility}
    \item[] Question: Does the paper fully disclose all the information needed to reproduce the main experimental results of the paper to the extent that it affects the main claims and/or conclusions of the paper (regardless of whether the code and data are provided or not)?
    \item[] Answer: \answerYes{} % Replace by \answerYes{}, \answerNo{}, or \answerNA{}.
    \item[] Justification: Broad details are provided in the main paper. The full details are included in Appendix. Code is available publicly.

\item {\bf Open access to data and code}
    \item[] Question: Does the paper provide open access to the data and code, with sufficient instructions to faithfully reproduce the main experimental results, as described in supplemental material?
    \item[] Answer: \answerYes{} % Replace by \answerYes{}, \answerNo{}, or \answerNA{}.
    \item[] Justification: The dataset has been shared. The code used for generating and evaluating the dataset was also shared, with instructions for creating environment and running the Python code.

\item {\bf Experimental setting/details}
    \item[] Question: Does the paper specify all the training and test details (e.g., data splits, hyperparameters, how they were chosen, type of optimizer) necessary to understand the results?
    \item[] Answer: \answerYes{} % Replace by \answerYes{}, \answerNo{}, or \answerNA{}.
    \item[] Justification: Broad details are provided in the main paper. The full details are included in Appendix. Code is available publicly.

\item {\bf Experiment statistical significance}
    \item[] Question: Does the paper report error bars suitably and correctly defined or other appropriate information about the statistical significance of the experiments?
    \item[] Answer: \answerYes{} % Replace by \answerYes{}, \answerNo{}, or \answerNA{}.
    \item[] Justification: This information is in the Appendices.

\item {\bf Experiments compute resources}
    \item[] Question: For each experiment, does the paper provide sufficient information on the computer resources (type of compute workers, memory, time of execution) needed to reproduce the experiments?
    \item[] Answer: \answerYes{} % Replace by \answerYes{}, \answerNo{}, or \answerNA{}.
    \item[] Justification: Details are included in Appendix \ref{appDetails}.
    
\item {\bf Code of ethics}
    \item[] Question: Does the research conducted in the paper conform, in every respect, with the NeurIPS Code of Ethics \url{https://neurips.cc/public/EthicsGuidelines}?
    \item[] Answer: \answerYes{} % Replace by \answerYes{}, \answerNo{}, or \answerNA{}.
    \item[] Justification: The research conducted in the paper fully conforms with the Code of Ethics. Guidelines:

\item {\bf Broader impacts}
    \item[] Question: Does the paper discuss both potential positive societal impacts and negative societal impacts of the work performed?
    \item[] Answer: \answerNA{} % Replace by \answerYes{}, \answerNo{}, or \answerNA{}.
    \item[] Justification: We foresee no immediate scope for potential malicious or unintended uses, fairness considerations, privacy considerations, and security considerations.
    
\item {\bf Safeguards}
    \item[] Question: Does the paper describe safeguards that have been put in place for responsible release of data or models that have a high risk for misuse (e.g., pre-trained language models, image generators, or scraped datasets)?
    \item[] Answer: \answerNA{} % Replace by \answerYes{}, \answerNo{}, or \answerNA{}.
    \item[] Justification: No pretrained language models, image generators, or scraped datasets are created.

\item {\bf Licenses for existing assets}
    \item[] Question: Are the creators or original owners of assets (e.g., code, data, models), used in the paper, properly credited and are the license and terms of use explicitly mentioned and properly respected?
    \item[] Answer: \answerYes{} % Replace by \answerYes{}, \answerNo{}, or \answerNA{}.
    \item[] Justification: The NoRA benchmark is used fully in line with its intended purpose. The corresponding paper has been cited. Other external resources include [Potassco / clingo] (https://potassco.org/), [google-deepmind/funsearch](https://github.com/google-deepmind/funsearch).  These resources are compatible with non-commercial research use, as we have cited them appropriately here. 

\item {\bf New assets}
    \item[] Question: Are new assets introduced in the paper well documented and is the documentation provided alongside the assets?
    \item[] Answer: \answerYes{} % Replace by \answerYes{}, \answerNo{}, or \answerNA{}.
    \item[] Justification: The main assets introduced in the paper are the learned samplers, which have been made available.

\item {\bf Crowdsourcing and research with human subjects}
    \item[] Question: For crowdsourcing experiments and research with human subjects, does the paper include the full text of instructions given to participants and screenshots, if applicable, as well as details about compensation (if any)? 
    \item[] Answer: \answerNA{} % Replace by \answerYes{}, \answerNo{}, or \answerNA{}.
    \item[] Justification: The paper does not involve crowdsourcing nor research with human subjects.

\item {\bf Institutional review board (IRB) approvals or equivalent for research with human subjects}
    \item[] Question: Does the paper describe potential risks incurred by study participants, whether such risks were disclosed to the subjects, and whether Institutional Review Board (IRB) approvals (or an equivalent approval/review based on the requirements of your country or institution) were obtained?
    \item[] Answer: \answerNA{} % Replace by \answerYes{}, \answerNo{}, or \answerNA{}.
    \item[] Justification: The paper does not involve crowdsourcing nor research with human subjects.

\item {\bf Declaration of LLM usage}
    \item[] Question: Does the paper describe the usage of LLMs if it is an important, original, or non-standard component of the core methods in this research? Note that if the LLM is used only for writing, editing, or formatting purposes and does \emph{not} impact the core methodology, scientific rigor, or originality of the research, declaration is not required.
    %this research? 
    \item[] Answer: \answerYes{} % Replace by \answerYes{}, \answerNo{}, or \answerNA{}.
    \item[] Justification: The central aim of this paper is to use LLMs to learn samplers for evaluating reasoning models. The way in which LLMs are used is described in detail in Section \ref{secMethodology}.

\end{enumerate}

\end{document}